%% file: dynerf.tex
\documentclass[10pt,twocolumn,letterpaper]{article}

\usepackage{iccv}
\usepackage{times}
\usepackage{epsfig}
\usepackage{graphicx}
\usepackage{amsmath}
\usepackage{amssymb}
\usepackage{dsfont}
\usepackage{color}
\usepackage{comment}
\usepackage{subfig}
\usepackage{wrapfig}
\usepackage{nicefrac}
\usepackage{float}
\usepackage[normalem]{ulem}
\usepackage{enumitem}
\usepackage[final]{animate}
\usepackage[export]{adjustbox}

\usepackage[title]{appendix}

\usepackage{multirow}

\usepackage[pagebackref=true,breaklinks=true,letterpaper=true,colorlinks,bookmarks=false]{hyperref}
\usepackage{authblk}

\iccvfinalcopy 

\def\iccvPaperID{1399} 

\definecolor{darkgreen}{RGB}{30,150,30}
\definecolor{darkblue}{RGB}{0,0,127}
\definecolor{darkyellow}{RGB}{171,133,0}
\definecolor{darkred}{RGB}{180,20,20}
\definecolor{darkmagenta}{RGB}{127,0,127}
\definecolor{darkcyan}{RGB}{0,127,127}
\definecolor{chromeyellow}{rgb}{1.0, 0.65, 0.0}
\definecolor{amber}{rgb}{1.0, 0.75, 0.0}

\newif\ifdrafting
\draftingfalse 

\ifdrafting
  \newcommand{\OG} [1] {\textcolor{darkgreen}{[OG: #1]}}
  \newcommand{\CW} [1] {\textcolor{darkblue}{[CW: #1]}}
  \newcommand{\SL} [1] {\textcolor{darkmagenta}{[SL: #1]}}
  \newcommand{\BE} [1] {\textcolor{darkyellow}{[BE: #1]}}
  \newcommand{\TODO} [1] {{\color{darkcyan}{\bf [TODO: #1]}}}

\else
  \newcommand{\OG} [1] {}
  \newcommand{\CW} [1] {}
  \newcommand{\SL} [1] {}
  \newcommand{\BE} [1] {}
  \newcommand{\TODO} [1] {}
  
\fi

\begin{document}

\title{Neural Trajectory Fields for Dynamic Novel View Synthesis}

\makeatletter
\renewcommand\Authfont{\fontsize{11.5}{14.4}\selectfont}
\renewcommand\AB@affilsepx{\qquad \protect\Affilfont}
\makeatother
\author[1,2]{Chaoyang Wang}
\author[1]{Ben Eckart}
\author[2]{Simon Lucey}
\author[1]{Orazio Gallo}
\affil[1]{NVIDIA}
\affil[2]{Carnegie Mellon University}
\renewcommand*{\Authsep} { \ \ \ \ \  }%
\renewcommand*{\Authands}{ \ \ \ \ \  }%

\twocolumn[{%
      \vspace{-7mm}
      \renewcommand\twocolumn[1][]{#1}%
      \maketitle
      \begin{center}
        \centering
        \vspace*{-9mm}
        \captionsetup{type=figure}
        \subfloat{\animategraphics[trim={0 20 750 20},width=0.18\linewidth,autoplay,loop]{10}{figures/teaser_new/kid_frames/}{00000}{00030}}\hspace{1mm}
        \subfloat{\includegraphics[trim={0 0 0 0},clip,width=0.29\linewidth]{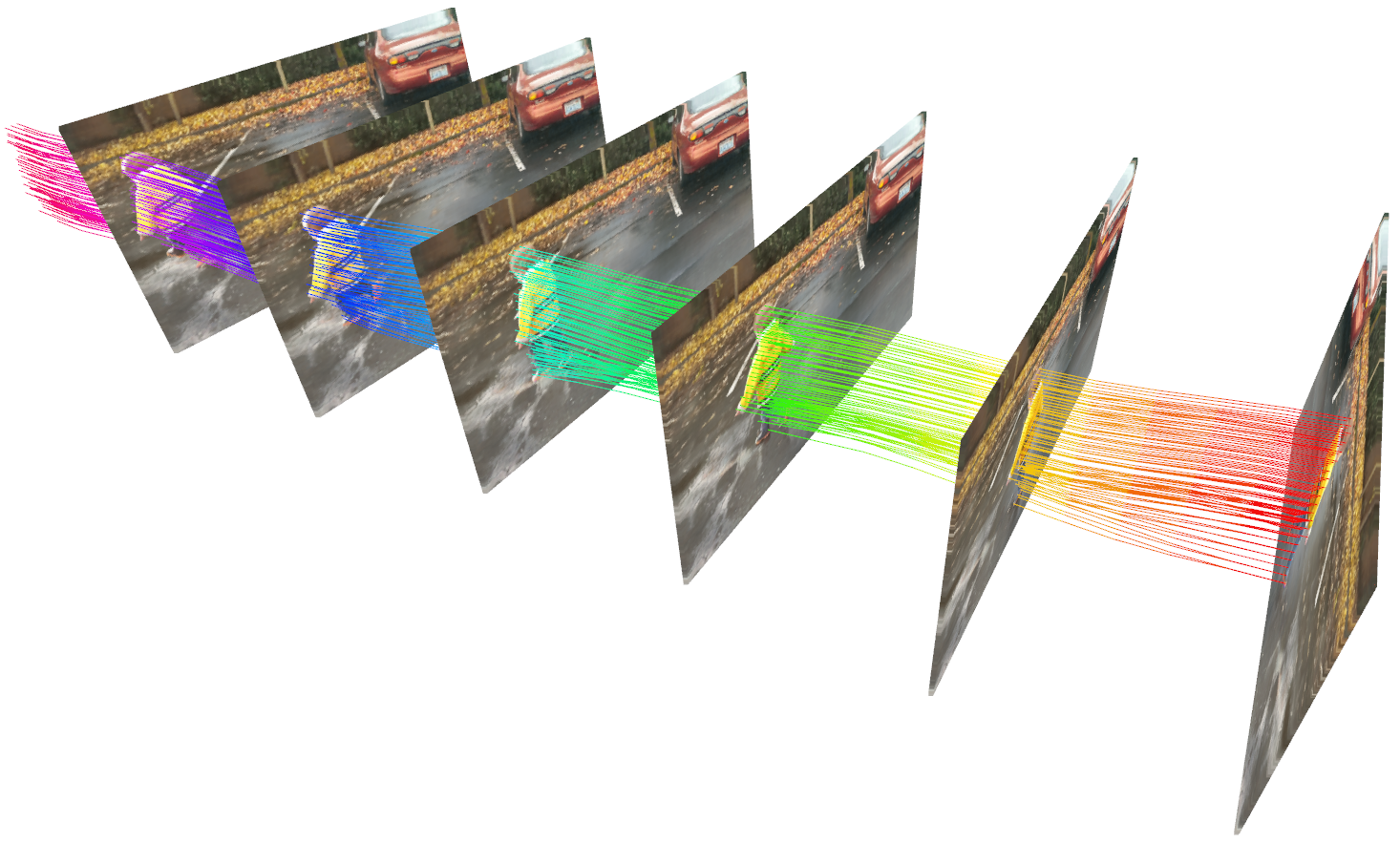}}
        \hspace{5mm}
        \subfloat{\animategraphics[trim={750 20 0 20},width=0.18\linewidth,autoplay,loop]{10}{figures/teaser_new/balloon_frames/}{00000}{00030}}\hspace{1mm}
        \subfloat{\includegraphics[trim={0 0 0 0},clip,width=0.27\linewidth]{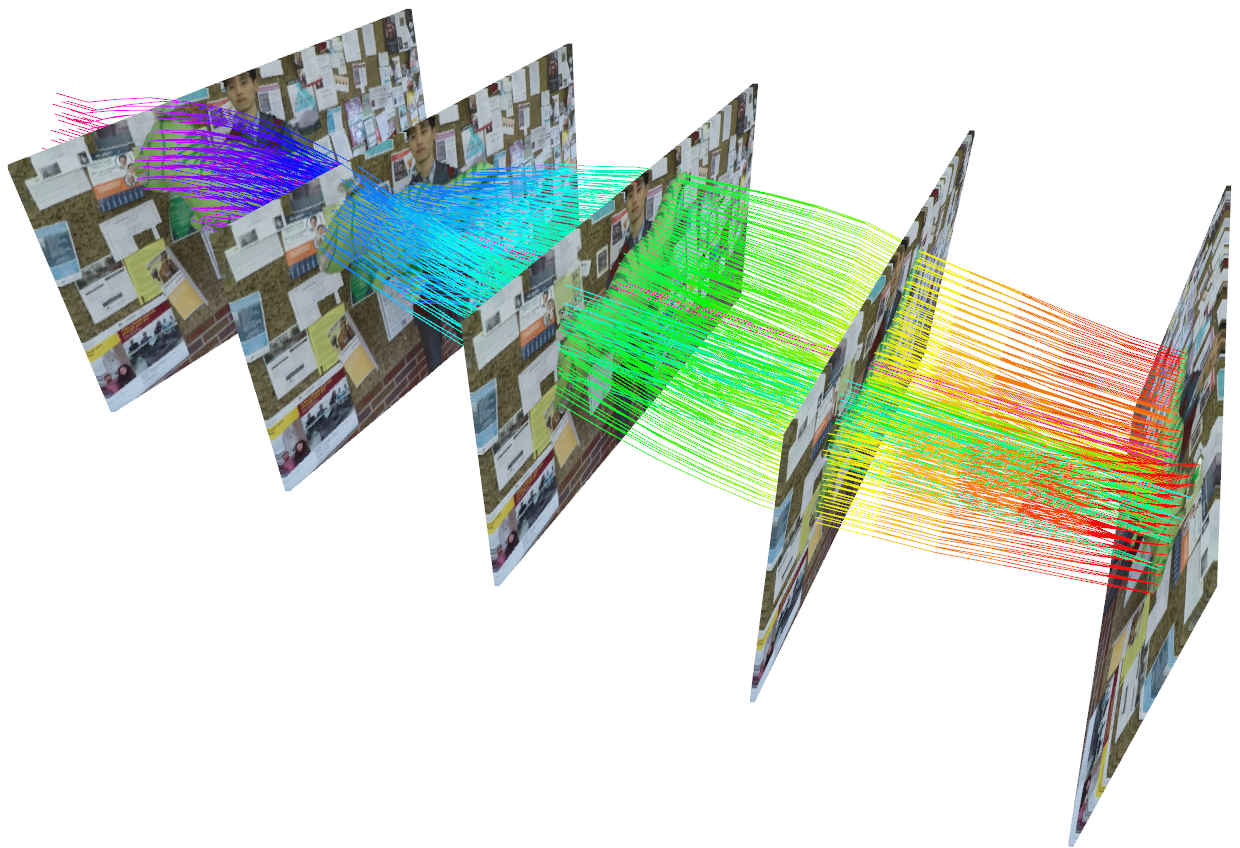}}
        \captionof{figure}{\small{Neural trajectory fields define dense, parametric, spacetime trajectories that enable consistent novel view synthesis even in the presence of dynamic and complex motion. The trajectories shown here are estimated for points in the first view. Note how faithfully they follow the dynamic objects. Parametrized using the Discrete Cosine Transform (DCT), these trajectories allow us to extrapolate motion even beyond the last available frame. \bf{To view the animations, please open this document in a media-enabled PDF viewer, such as Adobe Reader}.}
          \label{fig:teaser}}
      \end{center}%
    }]
\maketitle

\input{notations}

\begin{abstract}
  \input{abstract}
\end{abstract}

\section{Introduction}\label{sec:intro}
\input{intro}

\section{Related Work}\label{sec:related}
\input{related}

\section{Method}\label{sec:method}
\input{method_pt1}

\input{method_pt2}

\input{implementation}

\section{Evaluation and Results}\label{sec:results}
\input{results}

\section{Conclusions}\label{sec:conclusions}
\input{conclusions}


{\small
  \bibliographystyle{ieee_fullname}
  \bibliography{dynerf}
}
\newpage
\begin{appendices}
\input{supp}

\end{appendices}

\end{document}

%% file: notations.tex
\def\c{\mathbf{c}} 
\def\C{\mathbf{C}} 
\def\P{\mathbf{p}} 
\def\traj{\mathcal{T}} 
\def\d{\mathbf{d}} 
\def\coeff{\boldsymbol{\varphi}}

\def\p{\mathbf{p}} 

\def\cI{\mathcal{I}}
\def\cT{\mathcal{T}}
\def\p{\mathbf{p}}
\def\t{\mathbf{t}}
\def\R{\mathbf{R}}
\def\D{\mathbf{D}}
\def\A{\mathbf{A}}
\def\B{\mathbf{B}}
\def\I{\mathbf{I}}
\def\supi{{(i)}}
\def\bd{\mathbf{d}}
\def\bz{\mathbf{z}}
\def\bw{\mathbf{w}}
\def\M{\mathbf{M}}
\def\S{\mathbf{S}}
\def\W{\mathbf{W}}
\def\w{\mathbf{w}}
\def\x{\mathbf{x}}
\def\t{\mathbf{t}}
\def\cW{\mathcal{W}}
\def\cL{\mathcal{L}}
\def\bvarphi{\boldsymbol{\varphi}}
\def\btheta{\boldsymbol{\theta}}
\def\blambda{\boldsymbol{\lambda}}
\def\bPsi{\mathbf{\Psi}}
\def\bpsi{\boldsymbol{\psi}}

\def\s{\mathbf{s}}
\def\Real{\mathbb{R}}
\def\so{\mathfrak{so}}
\def\SO{\mathbb{SO}}
\def\1{\mathbf{1}}
\def\xy{\text{xy}}
\def\Z{\mathbf{Z}}
\def\d{\mathbf{d}}
\def\C{\mathcal{C}}
\def\tW{\widetilde{\W}}
\def\tD{\widetilde{\D}}
\def\td{\widetilde{\d}}
\def\tbPsi{\widetilde{\bPsi}}
\def\tP{\widetilde{P}}

\def\idx{{(i)}}

\newcommand{\SfM}{S\textit{f}M\xspace}
\newcommand{\SfC}{S\textit{f}C\xspace}
\def\SfMpp{S\textit{f}M++\xspace}

\newcommand{\centered}[1]{\begin{tabular}{l} #1 \end{tabular}}

%% file: abstract.tex
\vspace{-3mm}
Recent approaches to render photorealistic views from a limited set of photographs have pushed the boundaries of our interactions with pictures of static scenes.
The ability to recreate moments, that is, time-varying sequences, is perhaps an even more interesting scenario, but it remains largely unsolved.
We introduce DCT-NeRF, a coordinate-based neural representation for dynamic scenes.
DCT-NeRF learns smooth and stable trajectories over the input sequence for each point in space.
This allows us to enforce consistency between \emph{any two frames} in the sequence, which results in high quality reconstruction, particularly in dynamic regions.

%% file: intro.tex

\begin{figure*}[htbp]
    \centering
    \includegraphics[height=1.09in]{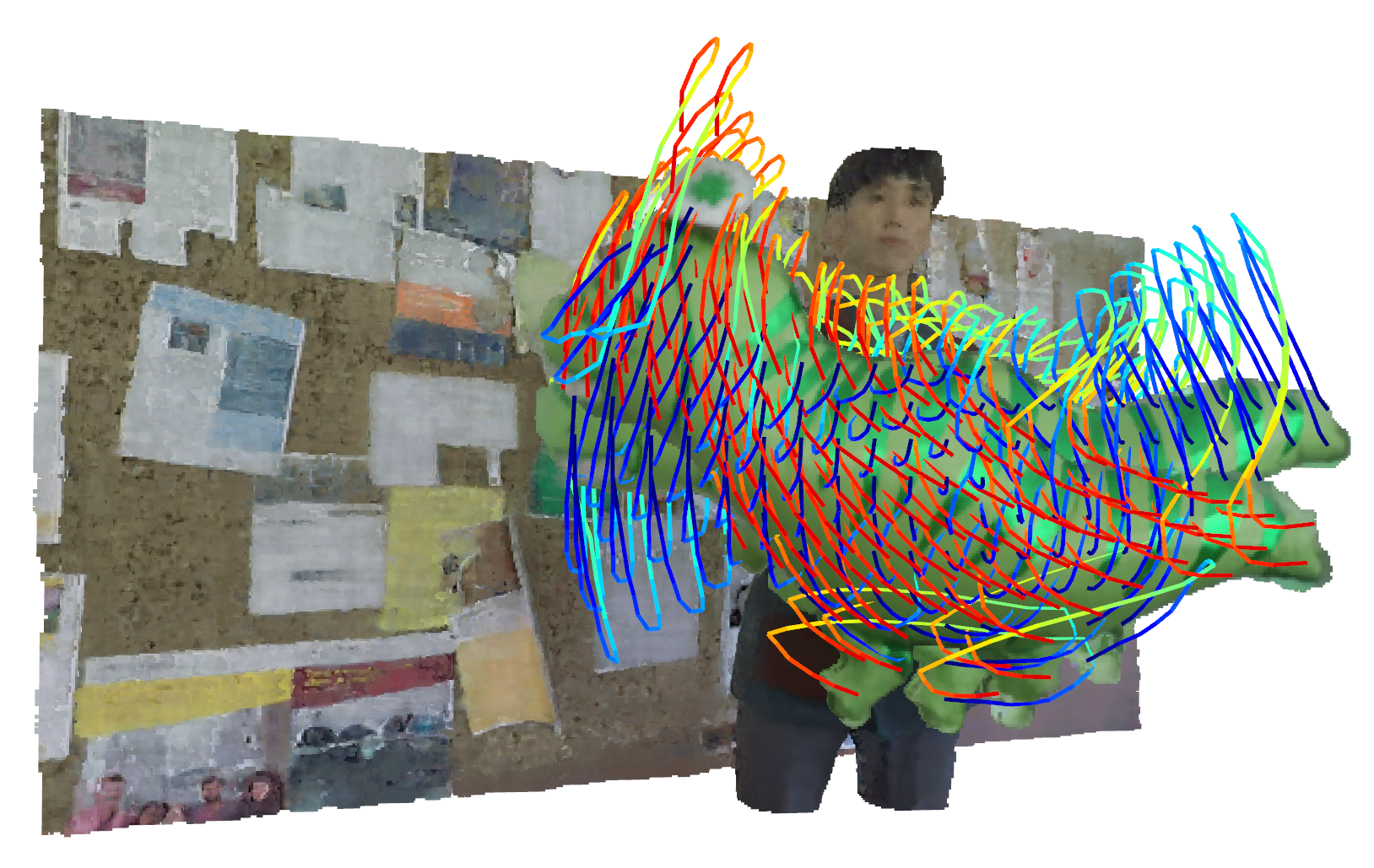}
    \includegraphics[height=1.09in]{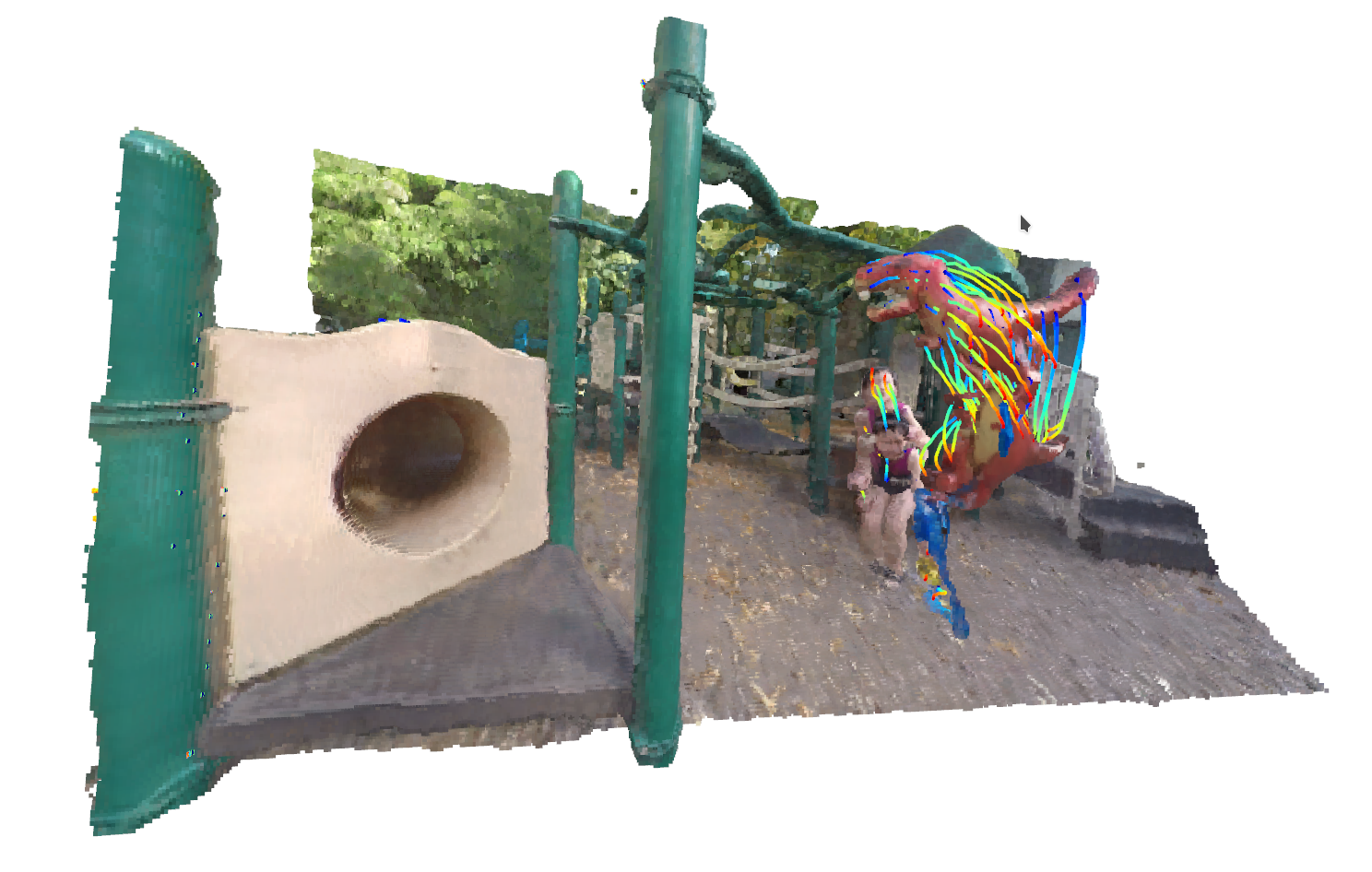}
    \includegraphics[height=0.9in]{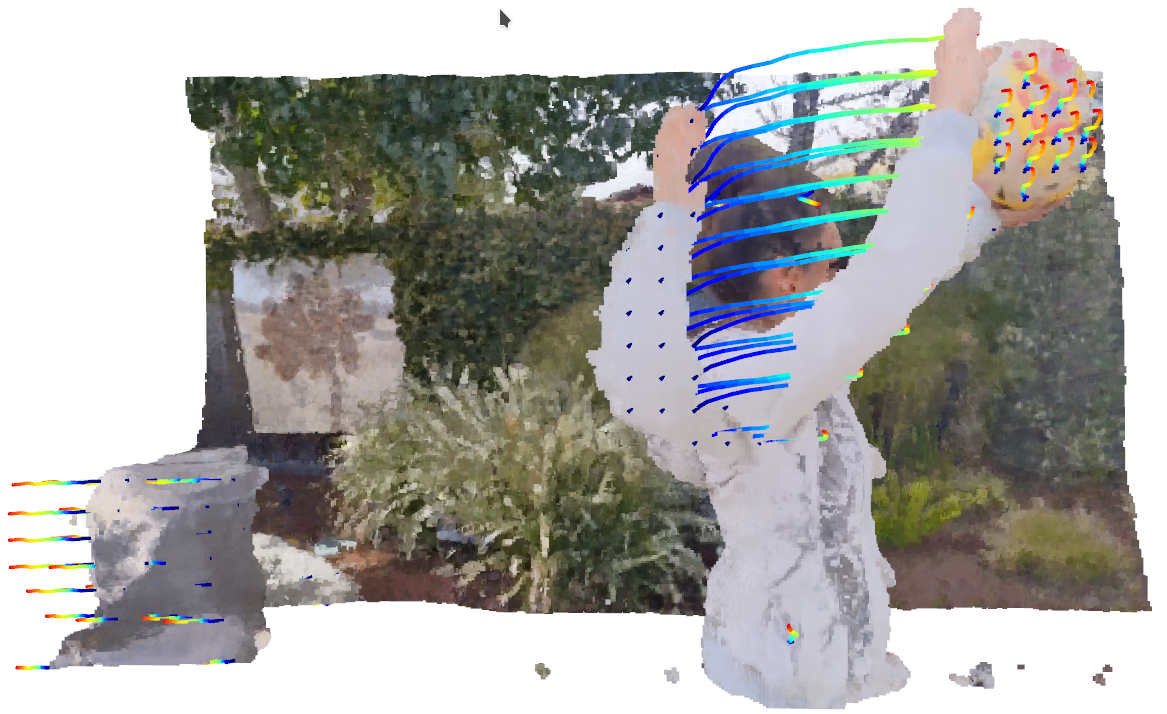}
    \includegraphics[height=1.09in]{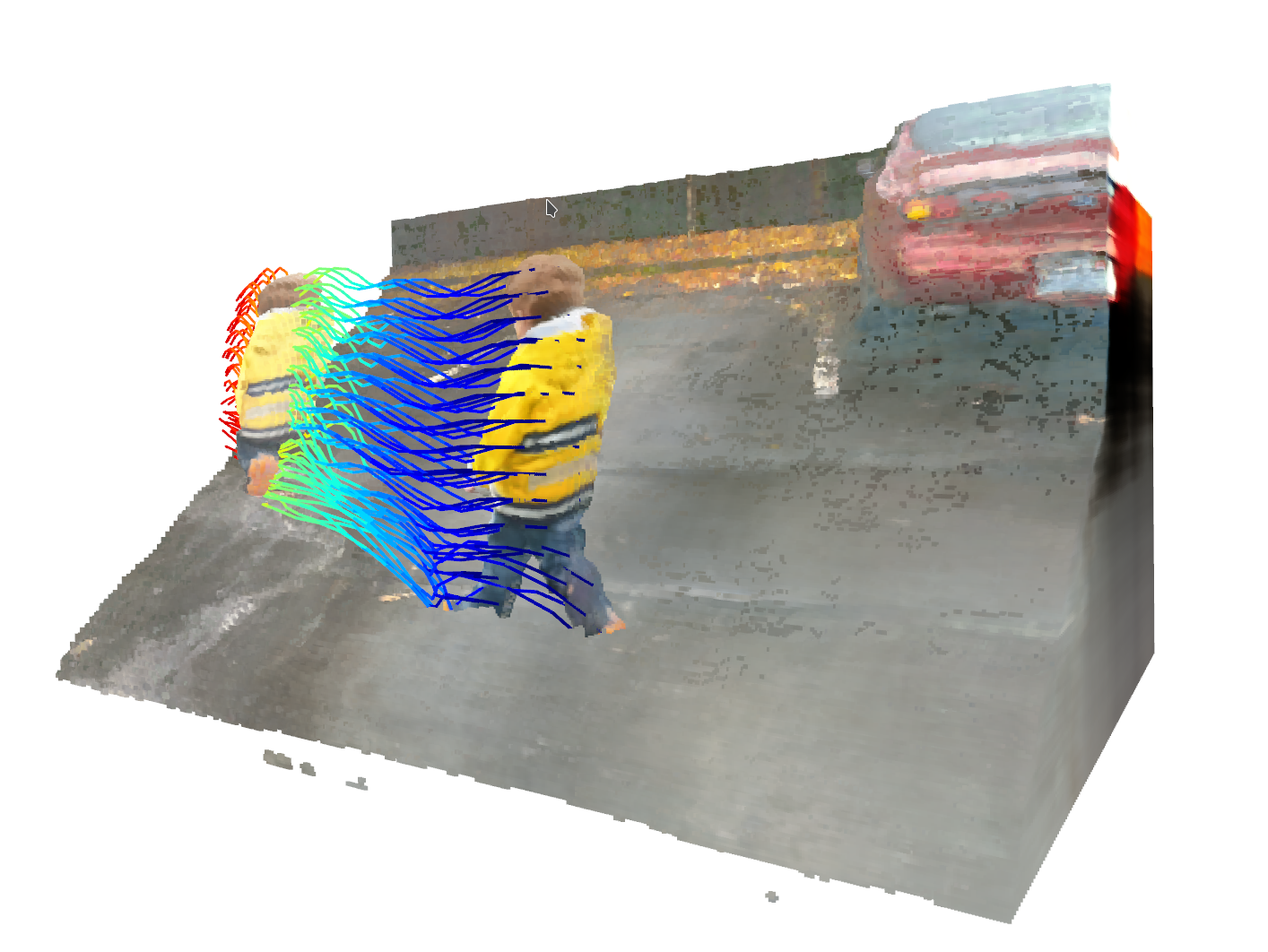}
    \caption{\small{Our method predicts dense sequence-long trajectories for each point in the scene. Here we show examples of trajectories sampled at sparse locations. The 3D scenes are rendered by overlapping the radiance fields from two different frames of each sequence. The visualized 3D trajectories are directly estimated from the output of the first radiance field.}}
    \label{fig:3Dtrajectories}
\end{figure*}

Novel view synthesis (NVS), the ability to create a photorealistic rendition of a scene from an unseen viewpoint, is a challenging, long-standing problem.
After impressive early attempts~\cite{chen1993view,levoy1996light,debevec1996modeling}, the computer vision and graphics communities made relatively slow progress towards a general and practical solution.
That changed with the recent exploration of neural rendering~\cite{tewari2020state}, in particular the Neural Radiance Field (NeRF)~\cite{Mildenhall20eccv_nerf}, a coordinate-based neural representation for differentiable volumetric rendering which has since fueled a new wave of interest~\cite{Liu20neurips_sparse_nerf,Schwarz20neurips_graf,Tewari20eurographics_neural_rendering}.

NeRF uses a multilayer perceptron (MLP), $\Psi$, to implicitly represent the geometry and radiance information of a \emph{static} 3D scene.
The color and transparency of a 3D point in space is estimated by querying the coordinates of the corresponding point and an associated viewing angle: $(\mathbf{c},\sigma)=\Psi(\P,\d)$. The network $\Psi$ is trained using a render-and-compare strategy, employing volumetric rendering from ray samples for view synthesis.
NeRF makes an assumption common to most existing methods: the subject must be static so that the epipolar geometry constraints hold across views.


Generalizing NeRF for dynamic scenes captured from a single camera is challenging, as it requires disambiguating displacements due to parallax from those due to non-rigid motion.
Recent pre-prints extend NeRF in one of two ways:
they either parametrize time explicitly, \ie, $(\mathbf{c}, \sigma)=\Psi(\P, t, \d)$~\cite{flow-fields,spacetime-nerf},
or they learn a warping field $\phi$ that maps the query coordinates at time $t$ to coordinates in a canonical frame, \ie, $(\mathbf{c},\sigma)=\Psi\left(\phi(\P, t), \d\right)$~\cite{nr-nerf,dnerf}.
When attempting to regularize the motion, these methods face a trade-off:
broadly speaking, the more powerful the prior (\eg, the subject is a portrait~\cite{portraitnerf}), the more stringent the assumptions on the scene content.
On the other hand, more general priors (\eg, piece-wise linear motion~\cite{flow-fields}) tend to be weaker in enforcing consistency in dynamic regions.

This tension is not unique to NeRF, or even to NVS methods.
A similar issue affects structure from motion (\SfM) estimation.
In that context, Akhter~\etal argue that motion lives in space and time and can be tackled by using priors in either domain~\cite{akhter2010trajectory}.
Motion, then, can be accounted for by using a combination of shape basis (space), which tend to be object-specific, or by using physical constraints on the trajectory (time), which are more object-agnostic~\cite{valmadre2012general,zhu2013convolutional,kumar2017spatio}.
Akhter~\etal posit that using trajectories parameterized as a linear combination of sinusoidal bases (\eg, DCT transform) offers a good compromise between motion regularization and generality of the priors.

Inspired by this observation, we propose DCT-NeRF, a new spatiotemporal, coordinate-based neural scene representation.
In addition to color and transparency, DCT-NeRF outputs continuous 3D trajectories across the whole input sequence: $(\traj, \mathbf{c}, \sigma)=\Psi(\P, t, \d)$, where $\traj$ is the discrete cosine transform trajectory of the point with spatiotemporal position $(\P,t)$.
As can be seen in Figure~\ref{fig:teaser}, this representation produces physically plausible motions that stay consistent across all frames, rather than simply predicting warping fields between neighboring frames.
Moreover, in contrast to existing works, the trajectory field provides arbitrarily long spatiotemporal correspondence predictions that allow us to establish a principled regularization scheme.

Rendering views from significantly different times with our trajectory parametrization requires careful handling of two factors.
First, we need to account for the potential change in reflectance of a point along its trajectory in both space and time.
We do this by allowing the predicted color of a spacetime location to be a function of time.
Second, we need to deal with regions that become occluded due to motion.
Li~\etal rely on a disocclusion mask learned for each pair of neighboring input images~\cite{flow-fields}, but this approach does not scale when comparing any two frames in the sequence---the number of masks is combinatorial with respect to the sequence length.
However, we note these problematic occlusion happen only when empty space at one time becomes occupied at another.
This observation allows us to discard regions that may get occluded by dynamic objects simply by leveraging the transparency field $\sigma$, which our network already estimates for volumetric rendering.

Figure~\ref{fig:3Dtrajectories} shows examples of reconstructed radiance fields and trajectories.
Note that the visualized trajectories are estimated by querying the network \emph{only at the time of the first frame}, but capture the motion of objects over the entire sequence of frames. Our method produces high-quality renderings of dynamic scenes from novel viewpoints, both numerically and visually.

%% file: related.tex

\begin{figure*}
    \centering
    \subfloat[Method Overview]{\includegraphics[height=1.3in]{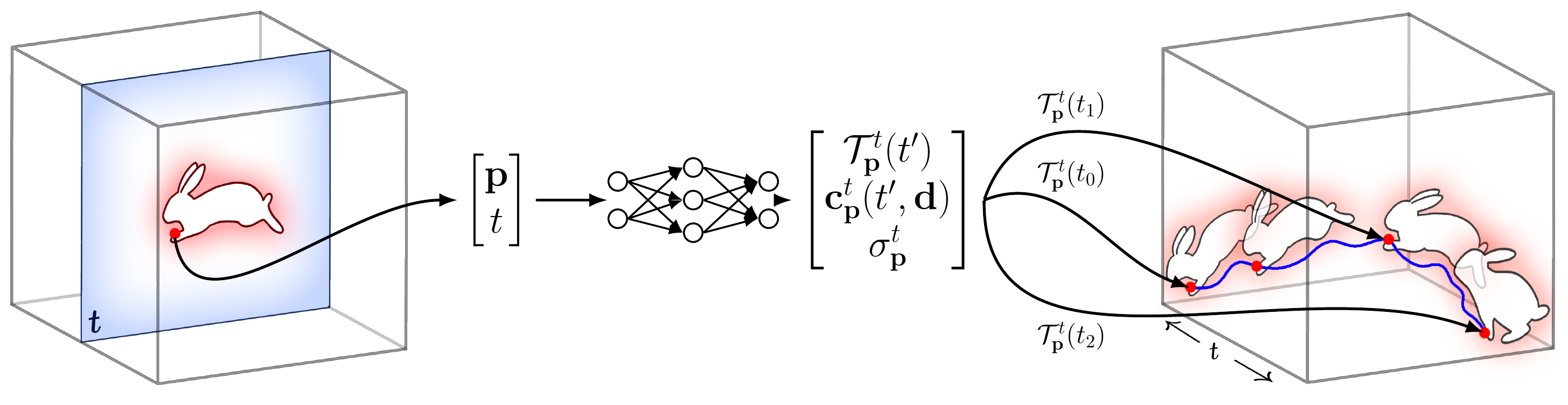}}\hspace{7mm}
    \subfloat[Types of Warping]{\includegraphics[height=1.3in]{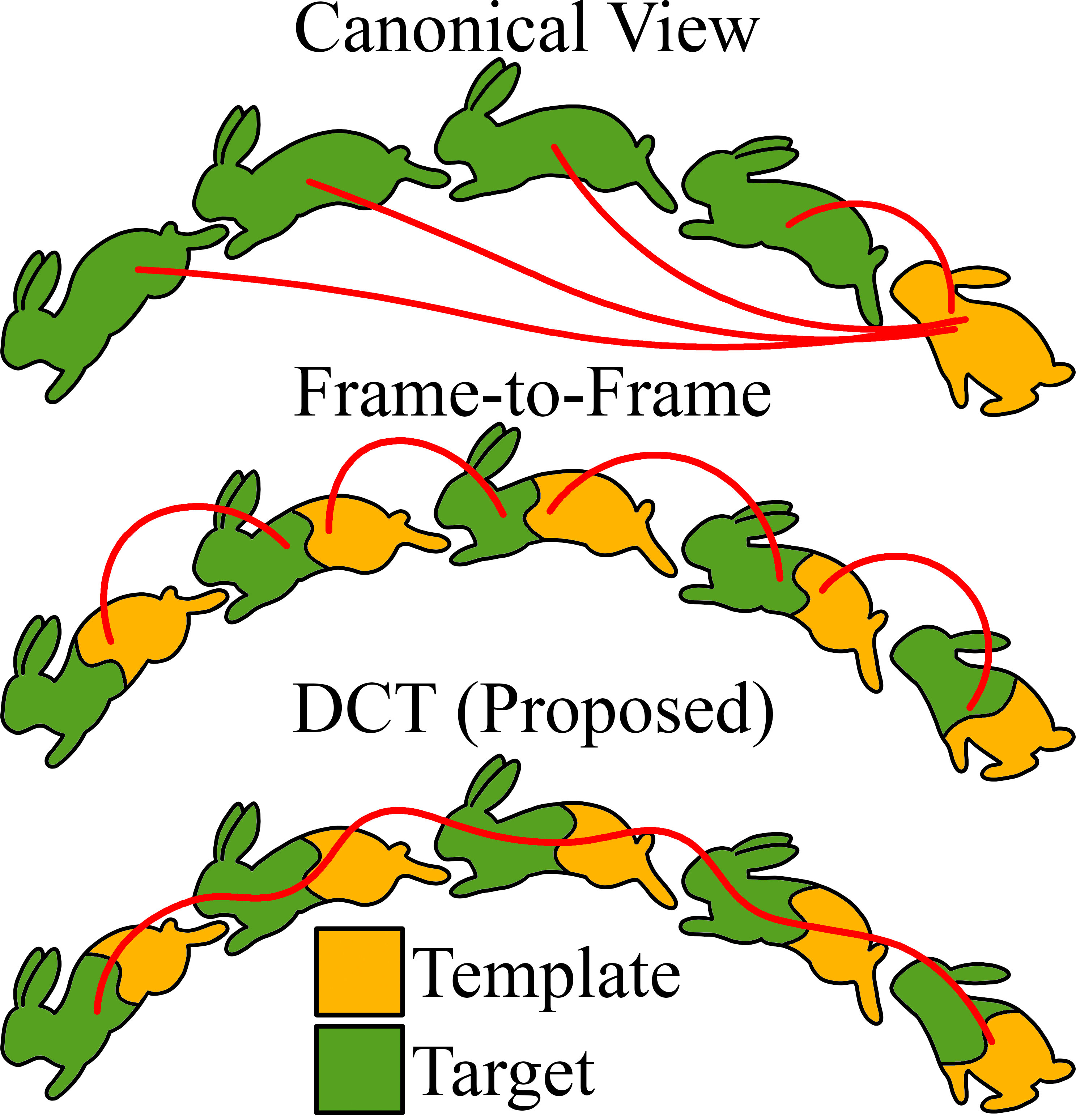}}
    \vspace{-3mm}
    \caption{\small{\textbf{(a)} We estimate the transparency $\sigma_\P^t$ and color $\c_\P^t$ for each spacetime location $(\P,t)$ with viewing direction $\d$. $\c_\P^t$ is modeled as a temporal-view dependent function. To model the deformation of dynamic objects, we also predict the trajectory $\traj_\P^t(\cdot)$ of any $(\P, t)$, which allows us to follow the point as it moves through space and time. We parametrize $\traj_\P^t$ as a DCT trajectory. \textbf{(b)} Existing approaches either define a canonical volume and warp all the frames to that one, or only warp neighboring frames. Thanks to our trajectories, our method can warp information from any frame to any frame.}}
    \label{fig:architecture}
\end{figure*}

\noindent\textbf{Novel view synthesis.}
Early attempts at rendering a scene from an unseen view point date back several decades~\cite{greene1986environment,chen1993view,levoy1996light,buehler2001unstructured,debevec1996modeling}.
In part due to novel deep learning techniques, we have observed a proliferation of methods in just the last few years, which can be categorized based on the way they represent the scene.

A traditional approach is to estimate depth explicitly and use it to warp pixels, or learned features, from the input images to the novel view~\cite{kalantari2016learning,penner2017soft,extremeview,stableview}.
Such scene representations offer flexibility, but inherit the shortcomings of depth-estimation algorithms, and require ways to explicitly deal with disocclusions at rendering time.
Thanks to advances in single-image depth estimation, novel views can now even be generated from just a single image~\cite{synsin}.

A second class of methods uses multiplane images (MPI)~\cite{zhou2018stereo}.
These algorithms learn a representation in which objects in the scene are associated to fronto-parallel layers.
At inference time MPIs are warped and fused into a novel view.
This approach implicitly regularizes depth and optimizes the layers to render the novel view.
As a result, MPI-based methods produce impressive NVS results~\cite{zhou2018stereo,pushingmpi,deepview}, even when using a single image as an input~\cite{singleMPI}, or a single semantic map instead of an RGB image~\cite{generativeVS,semanticsynth}.

The third and perhaps most promising strategy is to build an implicit neural representation of the scene~\cite{occnet,implicitnet}.
In their pioneering work, Mildenhall~\etal propose to train a multilayer perceptron (MLP) 
that can predict color and transparency for each point in space given a viewing direction~\cite{Mildenhall20eccv_nerf}.
In just over a year since its publication, NeRF has inspired a number of follow-up works that tackle issues such as improved quality~\cite{nerfpp}, performance~\cite{learnedinit,autoint}, or compositionality~\cite{derf,giraffe}. 

\noindent\textbf{Dynamic NVS.}
The problem of NVS is even more challenging for dynamic scenes because input views across time may capture differing scene geometry.
A dynamic scene can be ``frozen'' into a static one by using synchronized cameras~\cite{Zitnick2004HighqualityVV}.
Single-image depth estimation also effectively allows one to freeze a scene.
Luo~\etal use it to compensate for small motion~\cite{consistentdetph}, while Yoon~\etal use it to learn a full 4D representation of the scene, which they use to perform spatio-temporal image synthesis~\cite{globalcoherent}.

A number of pre-prints that augment NeRF for dynamic content have also recently appeared in parallel to our proposed research, which we report here for completeness.
Many make simplifying assumptions, such as the availability of a depth estimate~\cite{spacetime-nerf}, or that dynamic objects are distractors to remove~\cite{nerf-w}.
Other works target specific applications, such as self-portraits~\cite{nerfies,portraitnerf}.
Two closely related works are the works of Li~\etal, who use information from neighboring frames by learning a warping field~\cite{flow-fields}, and Tretschk~\etal who learn to warp input rays into a canonical volume~\cite{nr-nerf}.

The central idea for these methods, including ours, is that information must be warped between different time instants to combine information from multiple frames.
Both Tretschk~\etal~\cite{nr-nerf} and Gao~\etal~\cite{nerfies} define a canonical space-time volume and warp the information from all the frames to this template, either by bending the light rays~\cite{nr-nerf}, or by applying a warping field to the values from the non-canonical frames~\cite{nerfies}.
Instead of defining a canonical volume, for each frame in the sequence Li~\etal warp the information from the neighboring frames (\ie, $t\pm 1$, $t\pm 2$).
In a way these are opposite extremes: the former two have a single ``centralized'' representation of the scene, where all the input frames are mapped to the same volume, and the latter can only gather and leverage local information.
Our approach captures the advantages of both.
Because we use trajectories, we can estimate dense correspondences and combine information from any two frames in the sequence (see Section~\ref{sec:dct_traj}).
Figure~\ref{fig:architecture}(b) shows an illustration of the different strategies to combine dynamic information.

\noindent\textbf{Non-rigid structure from motion.}
NRSfM concerns the problem of recovering deforming 3D shapes using 2D point correspondences from multiple images. Bregler~\etal introduce this problem and propose the classical low-rank factorization approach\cite{bregler2000recovering}. Since then, two lines of work have emerged: shape-based and trajectory-based approaches. These two types of approaches are often complementary, and recent state-of-the-art employs both of them. Readers are referred to Jensen \etal~\cite{nrsfmbenchmark} for a comprehensive evaluation of NRSfM methods. We focus our discussion on trajectory-based approaches for their relevance to our paper.

Torresani~\etal utilize temporal information by imposing a linear motion model on top of a low-rank shape model, and show improvement on robustness against noise~\cite{torresani2008nonrigid}. Since then, different trajectory priors have been explored. The seminal work of Akhteret~\etal introduces a pure trajectory-based factorization method, which assumes that trajectories of different points in space can be represented as a linear combination of a small number of DCT trajectory bases~\cite{akhter2010trajectory}. Other methods have explored the convolutional structure of trajectories~\cite{zhu2013convolutional} as well as the assumption that different trajectories can be clustered into a small number of linear subspaces~\cite{kumar2017spatio}.

Taking inspiration from Akhteret~\etal, our approach represents continuous trajectories with DCT bases. A geometric analysis on how trajectories can be reconstructed using this representation is given by Park~\etal~\cite{park20103d}. One limitation to the low-rank approach of Akhteret~\etal is that the number of DCT bases needs to be tuned per sequence. Valmadre~\etal later argue that minimizing the response of high-pass filters (\eg, first- and second-difference filters) on trajectories eliminates the need to tune the number of bases, and also leads to more stable result for long sequences~\cite{valmadre2012general}. Following this insight, we minimize the difference between neighboring time steps on the trajectories without tuning the number of bases.

\noindent\textbf{Occupancy flow.} Niemeyer~\etal propose a temporally
and spatially continuous vector field to represent velocities~\cite{niemeyer2019occupancy}. To get the displacement of a point between different time steps, they need to integrate over the flow fields w.r.t. time, which leads to the number of queries of the network being proportional to time. In comparison, our trajectory field can output displacements between any time steps using only a single network query.

%% file: method_pt1.tex

We build on NeRF, an implicit neural representation that predicts the color $\c$ and the transparency $\sigma$ of any scene point, given a viewing direction~\cite{Mildenhall20eccv_nerf}: $(\c, \sigma) = \Psi(\P,\d)$.
NeRF estimates the color of a pixel in the desired view via volumetric rendering along the corresponding ray.
Given a number of posed views of a scene, NeRF can be trained by rendering pixels and comparing them with the corresponding ground-truth pixels.
This training strategy, however, requires the scene to be static, as a dynamic point $\P$ maps to different pixels in the input views.

To account for the point's motion, we propose to reconstruct its trajectory \emph{across the entire sequence}.
Inspired by classical works in the trajectory-based non-rigid structure from motion~\cite{akhter2010trajectory,valmadre2012general}, we parameterize the trajectory as a combination of sinusoidal bases, \ie, with a discrete cosine transform (DCT).
A few observations are in order.
First, a DCT trajectory is regular and smooth by construction, and has been shown to be a compact representation for the motion of deforming, dynamic objects~\cite{zhang2020we, mao2019}.
Second, all points in spacetime $(\P, t)\in\mathbb{R}^4$ are associated with a DCT trajectory $\traj_{\P}^{t}(\cdot)$,
and trajectories associated with any two spacetime points, \eg, $(\P_0,t_0)$ and $(\P_1,t_1)$, are expected to be the same if they correspond to the same physical point, \eg, the tip of the bunny's paw shown in Figure~\ref{fig:architecture}. This can be enforced by penalizing the difference between $\traj_{\P_0}^{t_0}$ and the trajectory associated with $(\traj_{\P_0}^{t_0}(t_1), t_1)$ for any $\P_0$, $t_0$, and $t_1$.
Finally, the ability to ``move'' the point in time following the trajectory allows us to compute the photometric error against the appropriate input pixels.

In the following we first formalize the description of our neural trajectory field, dubbed DCT-NeRF (see Section~\ref{sec:dctnerf}). We then show how the proposed representation is learned by enforcing photometric consistency through rendering with trajectories (see Section~\ref{sec:photo_loss}). The additional regularization terms we use are described in Section~\ref{sec:regularization}.
\subsection{DCT-NeRF}\label{sec:dctnerf}
Given a spacetime location $(\P,t)$, our neural field outputs the DCT coefficients (or frequency values) $\bvarphi$ for the trajectory, the embedding $\boldsymbol{\omega}$ for the temporal-view-dependent RGB color, and the transparency value $\sigma$:
\begin{equation}
    (\bvarphi_\P^t, \boldsymbol{\omega}_\P^t, \sigma_\P^t) = \Psi(\P, t),
\end{equation}
where superscript $t$ and subscript $\P$ are added to each output to denote that they are associated to the spacetime location $(\P, t)$.
The outputs $\bvarphi_\P^t$, $\boldsymbol{\omega}_\P^t$ are then used to condition the functional representation of the trajectory $\traj_\P^t: \mathbb{R}\rightarrow \mathbb{R}^3$ and the color output $\c_\P^t:\mathbb{R}\times\mathbb{R}^3\rightarrow\mathbb{R}^3$, \ie,
\begin{equation}
    \traj_\P^t(t') = f_{\text{DCT}^{-1}}(t', \bvarphi_\P^t),~~~~~\c_\P^t(t',\d) = f_\text{color}(t', \d, \boldsymbol{\omega}_\P^t),
\end{equation}
where $t'$ is the time input to the functions, $f_{\text{DCT}^{-1}}$ is the inverse DCT transform of the DCT coefficients $\bvarphi$ (details see Section~\ref{sec:dct_traj}), and $\c_\P^t$ is modeled as a neural network $f_\text{color}$.
The details of the network architecture are given in Section~\ref{sec:implementation}. We note that, unlike the original NeRF, our color output for a spacetime location $(\P, t)$ can vary over different time input $t'$ to account for specularities or illumination changes, even with fixed $\d$. This is crucial when enforcing photometric consistency across large temporal distance in Section~\ref{sec:temporal_photo}. In Figure~\ref{fig:radiance_anim}, we show animations of images rendered by $\c_\P^t(t',\d)$ with varying time input $t'$.

\subsubsection{DCT Trajectory Representation}\label{sec:dct_traj}
Given a sequence of $T$ frames, and the DCT coefficients $\boldsymbol{\varphi}\in\mathbb{R}^{3K}$ ($K<T$) associated with the trajectory $\traj=[x(t),y(t),z(t)]$, we can compute the motion along the $x$ axis as:
\begin{equation}\label{eq:idct_transform}
    x(t) = \sqrt{2/T}\sum_{k=1}^K \varphi_{x,k} \cos \left( \frac{\pi}{2T}(2t+1)k \right),
\end{equation}
where $\varphi_{x,k}$ denotes the $k$-th coefficient for the $x$ axis.
The $y$ and $z$ components of the trajectory have the same form.
Equation~\ref{eq:idct_transform} discards the $1^{\text{st}}$ component (\ie, $k=0$) of the inverse DCT transform.
We choose this formulation to remove the global offset, since we only care about the relative displacement between points on the trajectory, \ie $\Delta \P = \traj(t_1)-\traj(t_0)$.
In our experiments we set $K=T-1$ for short sequences. For longer sequences, we use smaller $K$ for computational efficiency. Note that reducing the number of coefficients is a classical approach to enforce smoothness of the trajectory, but it requires knowing the complexity of the motion as \emph{a priori}.

Training our network to directly predict the coefficients $\boldsymbol{\varphi}$ for each sampled spacetime location $(\P,t)$ offers a convenient mechanism to enforce a single trajectory for each object point: the coefficients $\boldsymbol{\varphi}$ predicted for the same point at time $t_0$ and at time $t_1$ should be the same.
Moreover, parameterizing a single trajectory for a point allows to predict the location of that point at any time $t$, whether that time instant was observed or not, see Figures~\ref{fig:teaser}~and~\ref{fig:3Dtrajectories}.
This enables enforcing photometric consistency between any two frames, as discussed below.

\subsection{Photometric Loss}
\label{sec:photo_loss}
\subsubsection{Frame-wise photometric consistency ($\mathcal{L}^{\text{photo}}_{t_0}$)}
Consider the color $\C^{t_0}$ of a pixel with ray direction $\d$ at time $t_0\in\{0,1,\cdots, T-1\}$.
The contribution of a point along the ray $\d$ at position $\P$ to the color $\C^{t_0}$ on the camera plane is
\begin{equation}\label{eq:contribution_from_t0}
    \delta\C_{\P}^{t_0}=\sigma_\P^{t_0} \c_\P^{t_0}(t_0, \d),
\end{equation}
where we make the dependence on $\P$, the sampled time $t_0$, and $\d$ explicit.
To render the corresponding pixel's color, we swipe all the points along a ray, and perform volumetric rendering:
\begin{equation}\label{eq:integral}
    \C^{t_0}=\int_{\xi_n}^{\xi_f} A(\xi)\delta\C_{\mathbf{o}+\xi\d}^{t_0}d\xi,
\end{equation}
where $A(\xi)=\exp\big(-\int_{\xi_n}^\xi \sigma_{\mathbf{o}+\nu\d}^{t_0}  d\nu\big)$ accounts for the attenuation along the ray, and $\mathbf{o}$ denotes the origin of the ray.\\
Given the color $\C^{t_0}_\text{GT}$ of a pixel from the input image at time $t_0$, we use the rendered color from Equation~\ref{eq:integral} to compute the photometric loss:
\begin{equation}\label{eq:standard_photo_loss}
    \mathcal{L}^{\text{photo}}_{t_0}= \left\| \C^{t_0} - \C_{\text{GT}}^{t_0} \right\|_2^2.
\end{equation}

\subsubsection{Temporal photometric consistency ($\mathcal{L}^{\text{photo}}_{(t_0;t_1)}$)}
\label{sec:temporal_photo}
Because we also estimate trajectories, we can integrate information from any other times, say time $t_1\in [0,T-1]$, in the same way that NeRF can compute the loss for any input frame \emph{despite the scene being dynamic}. First, given a spacetime location $(\P, t_0)$, we compute its point correspondence at time $t_1$ as $\P'=\P+\traj_{\P}^{t_0}(t_1)-\traj_{\P}^{t_0}(t_0)$. We can then follow Equation~\ref{eq:contribution_from_t0} and \ref{eq:integral} to render pixels at time $t_0$ using the warped radiance field from time $t_1$:
\begin{equation}\label{eq:contribution_from_t1}
    \delta\C_{\P}^{(t_0;t_1)}=\sigma_{\p'}^{t_1}\c_{\p'}^{t_1}(t_0, \d).
\end{equation}
The final color $\C^{(t_0;t_1)}$ is computed with volumetric rendering by integrating $\delta\C_{\P}^{(t_0;t_1)}$ as in Equation~\ref{eq:integral}. Temporal photometric consistency can then enforced by
\begin{equation}\label{eq:our_photo_loss}
    \mathcal{L}^{\text{photo}}_{(t_0;t_1)}= \left\| \C^{(t_0;t_1)} - \C_{\text{GT}}^{t_0} \right\|_2^2.
\end{equation}

\begin{figure}
    \vspace{-3mm}
    \input{figures/occlusion/pocc_figure}
    \vspace{-3mm}
    \caption{\small{Space that is empty at $t_0$ and becomes occupied at $t_1$ causes rendering issues (Rend. w/o $p_\text{occ}$). We predict the probability of a region to cause such issues ($p_\text{occ}$) and downweight their contribution to the rendering process accordingly (Rend. w/o $p_\text{occ}$).}}\label{fig:disocclusions}
\end{figure}
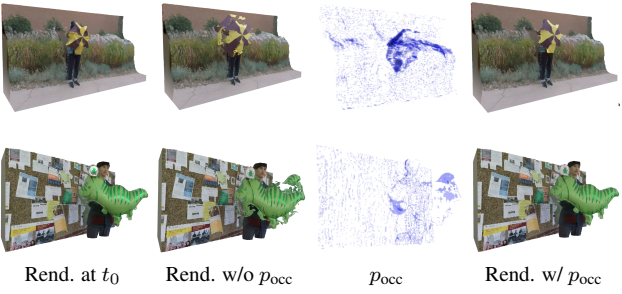

%% file: figures/occlusion/pocc_figure.tex

\newcommand{\insetwidth}{0.24\linewidth}

\newlength{\pocheight}
\settoheight{\pocheight}{\includegraphics[width=\insetwidth]{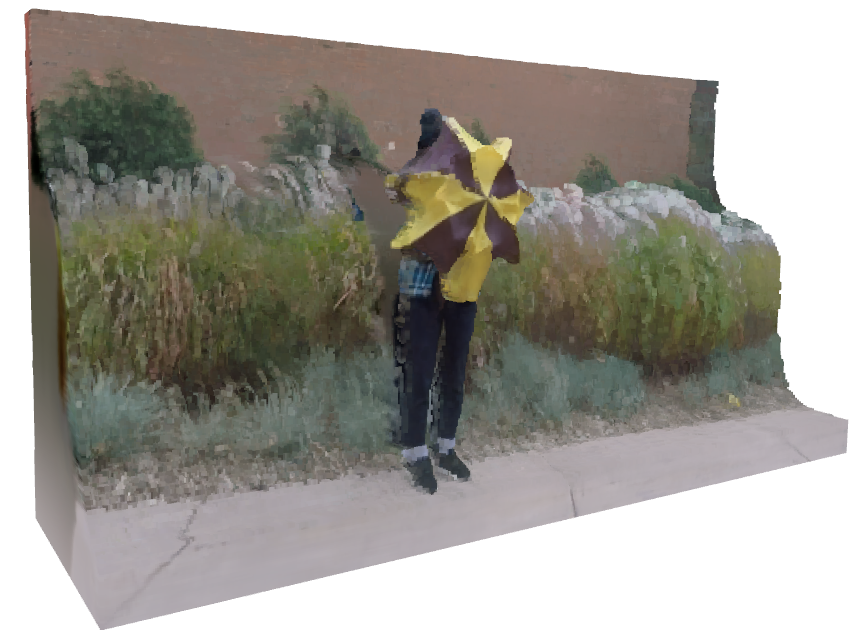}}
\centering
\captionsetup[subfigure]{labelformat=empty}
\subfloat{\includegraphics[width=\insetwidth]{figures/occlusion/ref_rf.png}}\
\subfloat{\includegraphics[width=\insetwidth]{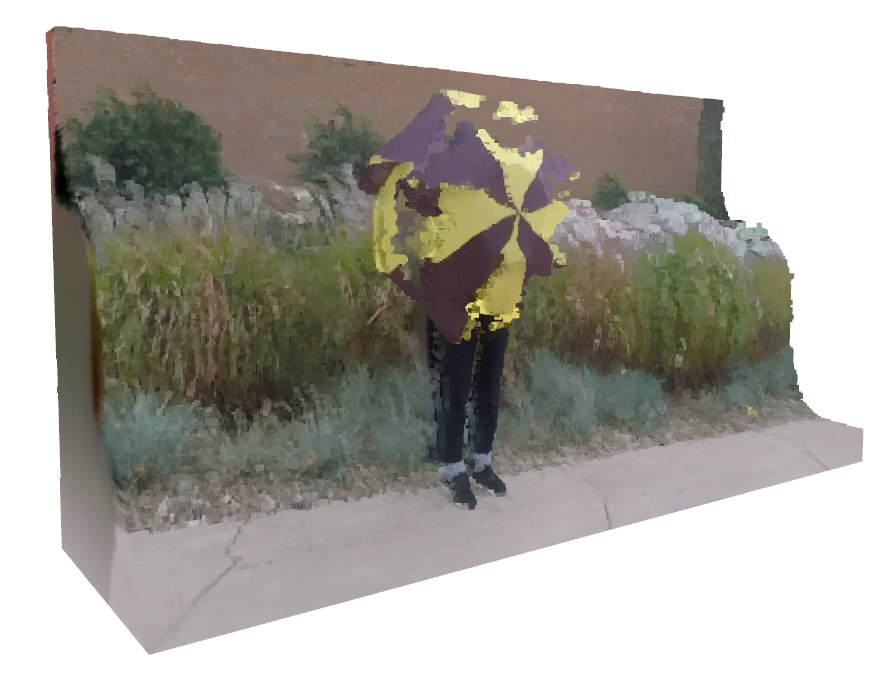}}\
\subfloat{\includegraphics[width=\insetwidth, height=\pocheight]{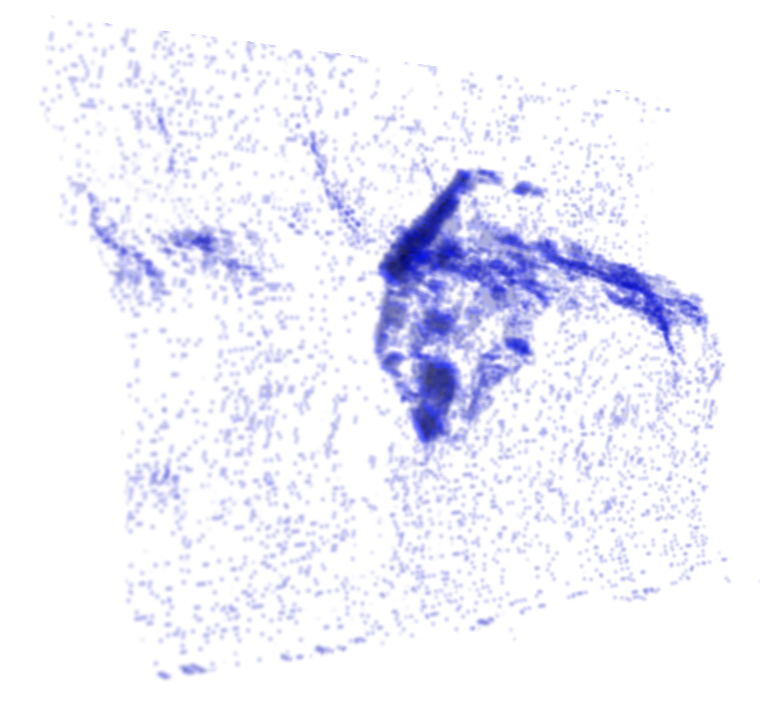}}\
\subfloat{\includegraphics[width=\insetwidth]{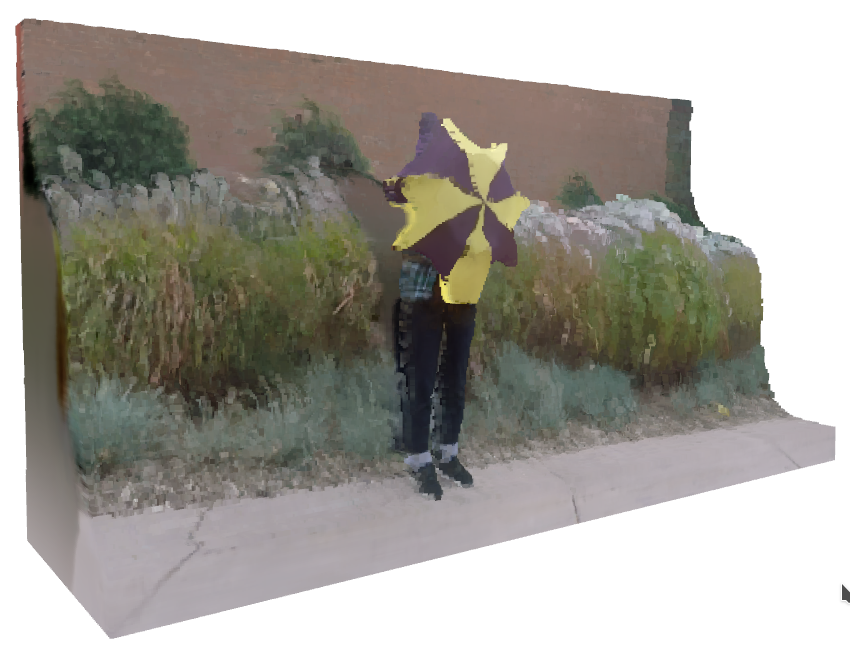}}\\
\subfloat[Rend. at $t_0$]{\includegraphics[width=\insetwidth]{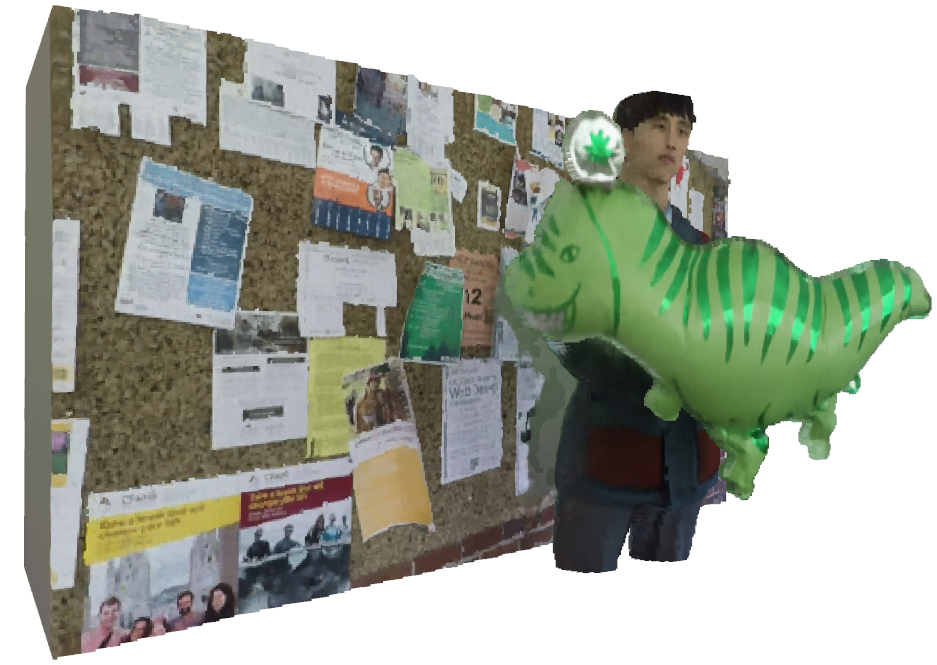}}\
\subfloat[Rend. w/o $p_\text{occ}$]{\includegraphics[width=\insetwidth]{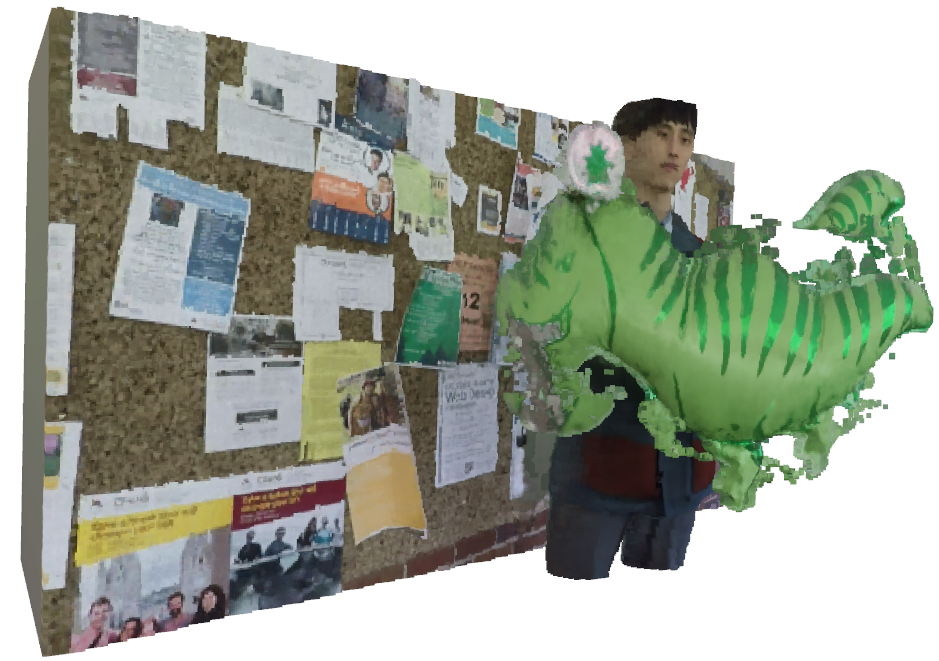}}\
\subfloat[$p_\text{occ}$]{\includegraphics[width=\insetwidth, height=\pocheight]{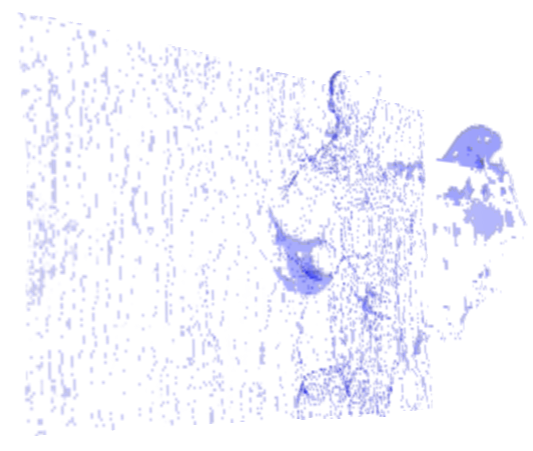}}\
\subfloat[Rend. w/ $p_\text{occ}$]{\includegraphics[width=\insetwidth]{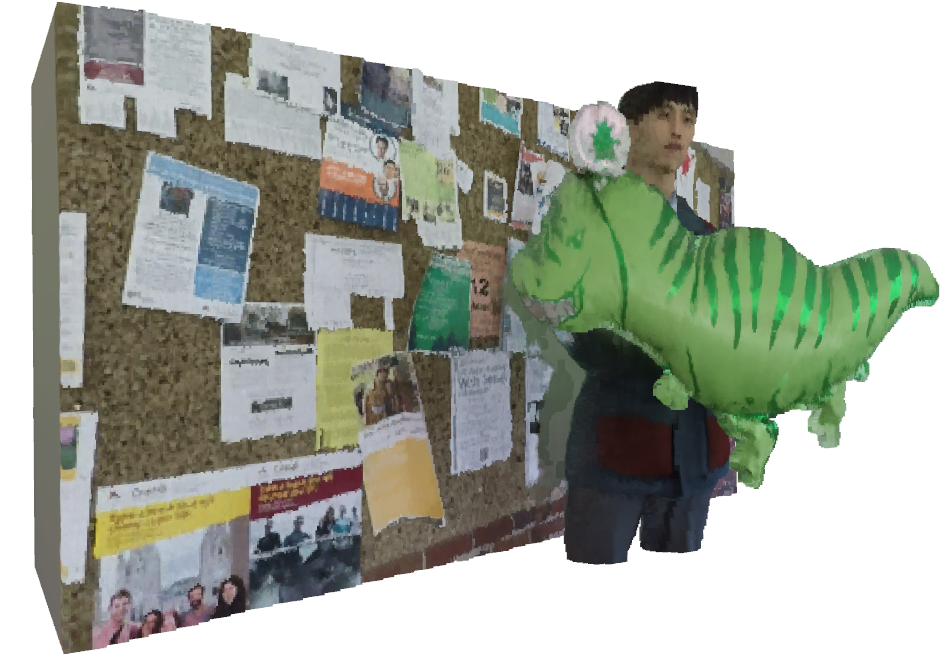}}\

%% file: method_pt2.tex

\paragraph{Handling Temporal Occlusions.}\label{sec:disocclusion}
Equation~\ref{eq:contribution_from_t1} does not account for temporal occlusions.
Note that these regions are different from the typical spatial occlusions induced by the camera displacement.
Consider a point $\P$ sampled in an empty area of the scene at time $t_0$, in other words $\sigma_\P^{t_0}=0$ and consequently $\delta\C^{t_0}_\P=0$. It is also expected that $\traj_\P^{t_0}(t_1)=0, \forall t_1$ since empty regions do not move, and thus $\p'=\p$.
Then suppose at some time $t_1$ an object moves and occupies position $\P$, which implies $\sigma_{\P'}^{t_1} = \sigma_\P^{t_1}\neq 0$.
Applying Equation~\ref{eq:contribution_from_t1} to this scenario yields

\begin{equation}
    \delta\C^{(t_0; t_1)}_{\P} = \sigma_{\p}^{t_1}\c_{\p}^{t_1}(t_0, \d) \approx \delta \C_\p^{t_1} \ncong \delta \C_\p^{t_0}.
\end{equation}

That is, even though $(\P,t_0)$'s trajectory is correctly estimated (zero motion), we use the wrong color when integrating Equation~\ref{eq:integral}.
One could identify and exclude these regions by learning a mask, as suggested by Li~\etal~\cite{flow-fields}.
However, comparing any two time stamps in the input sequence, as our method does, requires a combinatorial number of such maps.
We observe that, barring interpenetration of objects, and neglecting objects that only appear in a subset of frames, this issue only manifests in space that is empty at $t_0$ and becomes occupied at another time $t_1$.
Conveniently, our network predicts $\sigma$, which we use to estimate a probability of ``empty-ness'':
\begin{equation}\label{eq:disocclusions_weights}
    p_\text{e}(\P,t)=\text{sigmoid}(-k_1\sigma_\P^{t}+k_2),
\end{equation}
where we empirically set $k_1=2$, and $k_2=3$. Treating $k_1$ and $k_2$ as learnable parameters results in equal or degraded quality.
The probability that empty space at $(\P,t_0)$ becomes occluded at time $t_1$ can then be estimated as
\begin{equation}\label{eq:p_occ}
    p_\text{occ}(\P, t_0, t_1)=p_\text{e}(\P,t_0)\big(1-p_{e}(\P',t_1)\big),
\end{equation}
which we can use to modify Equation~\ref{eq:contribution_from_t1} to downweigh the contribution of occluded regions:
\begin{multline}
    \delta\C^{(t_0; t_1)}_{\P} = p_\text{occ}(\P, t_0, t_1) \cdot \delta\C^{t_0}_{\P} + \\
    (1-p_\text{occ}(\P, t_0, t_1))\cdot\sigma_{\p'}^{t_1}\c_{\p'}^{t_1}(t_0, \d).
\end{multline}
Figure~\ref{fig:disocclusions} shows a rendering of a frame using the warped radiance field from a different time, with and without $p_\text{occ}$ to handle occlusions.
It also shows the probability $p_\text{occ}$ over the whole space.

\subsection{Regularization}\label{sec:regularization}
To encourage our system to converge to the correct solution,
we follow the common practice of adding regularizers to our optimization.

\noindent\textbf{Cycle consistency ($\mathcal{L}^{\text{cycle}}$).} To enforce the network to make temporally consistent estimates, we add an $L_1$ loss on the difference of $(\traj,\c,\sigma)$ for corresponding points, where $\traj$ is represented by the DCT coefficients and the value of $\c$ is queried with the same $(\d,t)$ input.
To account for disocclusions, we weigh $\mathcal{L}^{\text{cycle}}$ by $1-p_\text{occ}$.

\noindent\textbf{Single visible surface ($\mathcal{L}^\text{SVS}$).} We want to encourage the source of radiance to be concentrated on a single surface area visible to the current view point.
This means that most of the energy of the attenuation coefficient $A(\xi)$ in Equation~\ref{eq:integral} along the ray should be concentrated in a small window of size $r$:
\begin{equation}
    \mathcal{L}^{\text{SVS}} = 1 - \max_{z_0} \int_{z=z_0}^{z=z_0+r} \tilde{A}(z)dz,
\end{equation}
where $\tilde{A}$ denotes that $A$ is normalized to sum to one along the ray.

\noindent\textbf{Trajectory regularizers ($\mathcal{L}^\text{traj}$).} Our trajectory field is regularized by the combination of the following terms: (i) the spatial smoothness term, which calculates the $L_1$ norm of the spatial derivative of scene flow;
(ii) an as-rigid-as-possible deformation loss~\cite{kumar2017monocular}, which is used to encourage the local deformation in occupied regions (measured by $p_\text{e}$) to be isometric; (iii) the temporal smoothness term which penalizes the $L_1$ norm of the scene flow.

\noindent\textbf{Depth \& flow.} We use precomputed optical flow~\cite{raft} to provide local supervision for the projection of $\traj$ between neighboring frames. We also employ single-image depth estimation~\cite{midas} to regularize $\sigma$ with a loss invariant to affine transformations~\cite{midas,wsvd}.

\begin{figure*}
    \includegraphics[width=\textwidth]{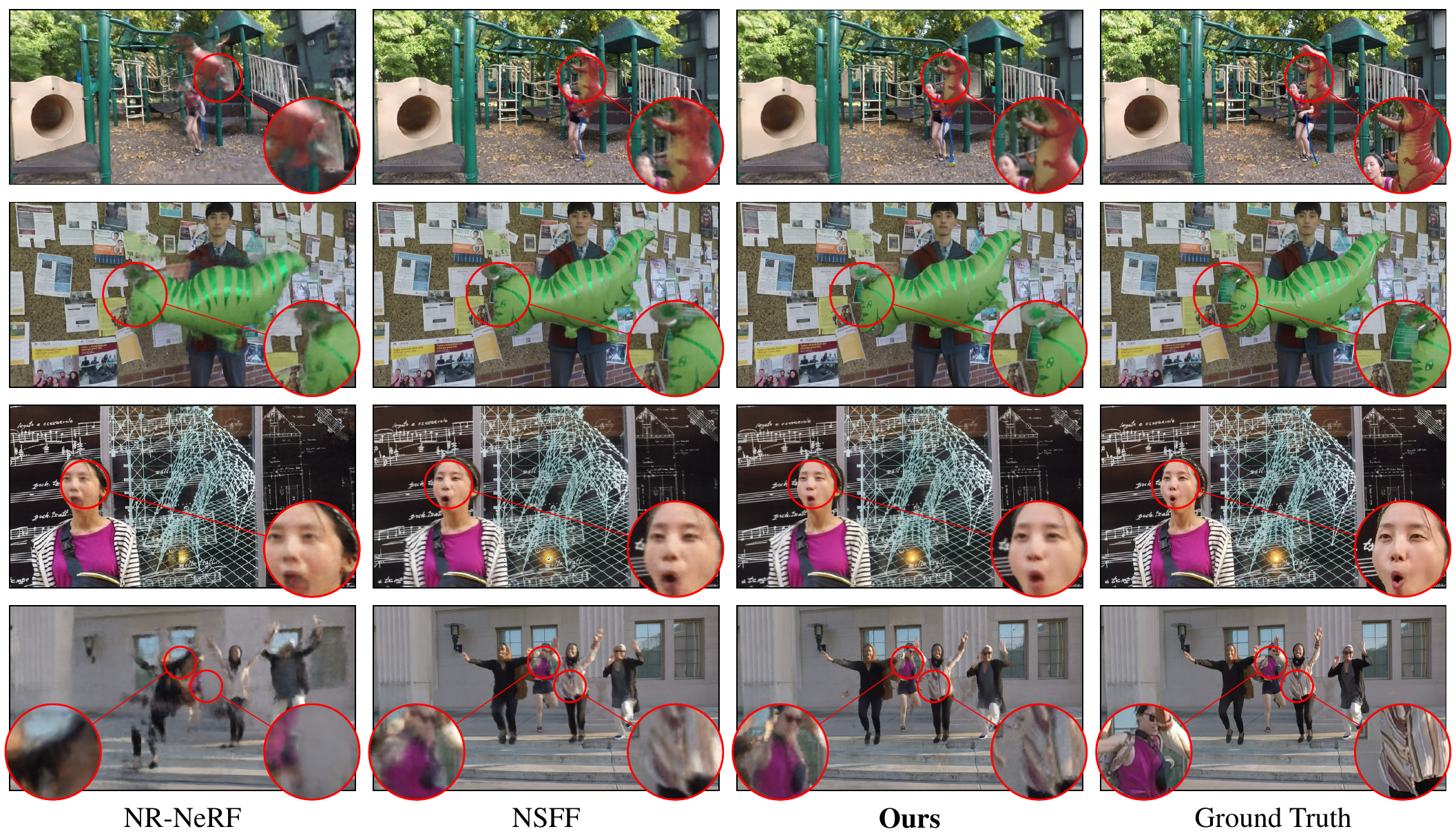}
    \vspace{-5mm}
    \caption{\small{Comparisons. Our trajectory-based representation allows our method to perform well---particularly in dynamic regions.}}\label{fig:results_with_insets}
\end{figure*}

%% file: implementation.tex

\subsection{Implementation}\label{sec:implementation}
\noindent\textbf{Optimization objective.} Given a pixel sampled at time $t_0$, we compute the loss as a combination of photometric losses and regularization terms. The photometric loss is evaluated as the sum of $\mathcal{L}^\text{photo}_{t_0} + \mathcal{L}^\text{photo}_{(t_0;t_0\pm 1)} + \mathcal{L}^\text{photo}_{(t_0;t_1)}$, which enforces the rendering consistency between neighboring time steps, as well as a randomly sampled time step $t_1$. We employ a local-to-global strategy to sample $t_1$.
That is, during the initial optimization iterations, $t_1$ is limited to $[t_0-2, t_0+2]$.
We then exponentially increase the temporal radius during optimization, until the sample range of $t_1$ covers the whole sequence.

We find that adjusting the weights for certain regularization terms during the different stages of the optimization is critical for their success.
First, when the weights for the depth and optical flow loss are not linearly decreased, as proposed by Li~$\etal$~\cite{flow-fields}, the network may be forced to overfit to the noise in the pre-computed depth and optical flow maps.
Second, gradually increasing the weight of $\mathcal{L}^\text{svs}$ from a small value, \eg, $1e-5$, helps to prevent getting stuck in poor local minima in the early stage of training. We use the same hyperparameters for all our experiments, and the details are included in the supplementary.

\noindent\textbf{Static background.} The consistency of the background across time is crucial for the viewing experience of dynamic NVS. Martin-Brualla~\etal show success in learning a rigid NeRF for static background from images with dynamic objects~\cite{nerf-w}. Their approach is to modify the rigid NeRF model to output an additional 3D blending field so as to exclude interference from the dynamic regions of input images. Similarly, Li~\etal~\cite{flow-fields} also include a rigid NeRF and blend the rigid and dynamic radiance fields to render an additional image when evaluating the photometric loss. We follow the approach of Li~\etal and observe an improvement in the stability of the background.

\noindent\textbf{Network architecture.} We adapt the architecture from the original NeRF~\cite{Mildenhall20eccv_nerf} implementation, with the following changes: (i) we concatenate an additional positional encoding of time to the input; (ii) parallel to the linear layers that predict $\sigma$, we add another linear layer to output the DCT coefficients; (iii) we concatenate the directional inputs with the positional encoding of time when calculating the color.

%% file: results.tex

\begin{figure}
	\centering
	\animategraphics[trim={0 0 0 0},width=\linewidth,autoplay,loop]{10}{figures/spacetime_radiance_anim/}{00000}{00023}
	\caption{\small{For photometric consistency across time in the presence of dynamic specularities and shadows, our approach allows the predicted radiance to be rendered with respect to any other reference time across the sequence. Given the leftmost column's time in the sequence as a reference, the rightmost two columns show two different points in spacetime but rendered using the predicted radiance values of the reference time. \bf{To view this animation, please open this document in a media-enabled PDF viewer, such as Adobe Reader.}}}\label{fig:radiance_anim}
\end{figure}

\noindent\textbf{Datasets.}
We use eight diverse scenes from the NVIDIA Dynamic Scene Dataset~\cite{globalcoherent} to validate our method with both numerical and visual comparisons. For each scene, the dataset provides 12 sequences captured by synchronized cameras at fixed positions. Then camera motion is synthesized by choosing frames from different cameras at each time step. We extract 24 frames per scene as input sequences for training, and report results on the held-out 11 images for each frame.

\noindent\textbf{Trajectories.}
Figure~\ref{fig:teaser} shows the trajectory estimated at the time of the first frame traced out over the whole duration of the sequence.
Notice that our trajectories are stable and smooth across the entire sequences, even when the motion changes direction, as is the case of the scene on the right.
The trajectories can be sampled as densely or as sparsely as needed.
\noindent
Figure~\ref{fig:3Dtrajectories} shows trajectories for the same sequences of Figure~\ref{fig:teaser} and two more overlaid to a 3D visualization of the radiance field computed by our method.

\noindent\textbf{Dynamic specularities and shadows.}
Because we allow the color $\c$ to vary with time,
we can render one frame with the radiance of another.
Figure~\ref{fig:radiance_anim} shows an example of this (please view the animation in a media-enabled reader).
In this experiment, we apply the radiance of a reference frame (the leftmost in the figure) to two different target frames, animated over the entire sequence. Notice how the dynamically changing specularities and shadows correctly correspond to the reference image.
This is possible because our method correctly estimates both radiance and trajectories over time.

\noindent\textbf{Ablation study.}
We analyze the impact of the regularization terms we describe in Section~\ref{sec:regularization} on the overall quality of the results. We use our model trained without the static background loss as a baseline.
Table~\ref{table:ablation} shows the numerical results when using only the neighboring frames for $\mathcal{L}^{\text{photo}}$ (``local only''), removing the occlusion weights ($p_{\text{occ}}$), and removing other regularization terms one at a time. Through our numerical analysis, as well as a thorough visual inspection of the results, we observe that these terms are statistically beneficial.
Removing any of them from our formulation introduces \emph{some} artifacts in \emph{some} of the results. 
However, they are not central to our results.
We refer the reader to the supplementary material for visual examples.

\begin{figure}
	\centering
	\includegraphics[width=1.02\columnwidth]{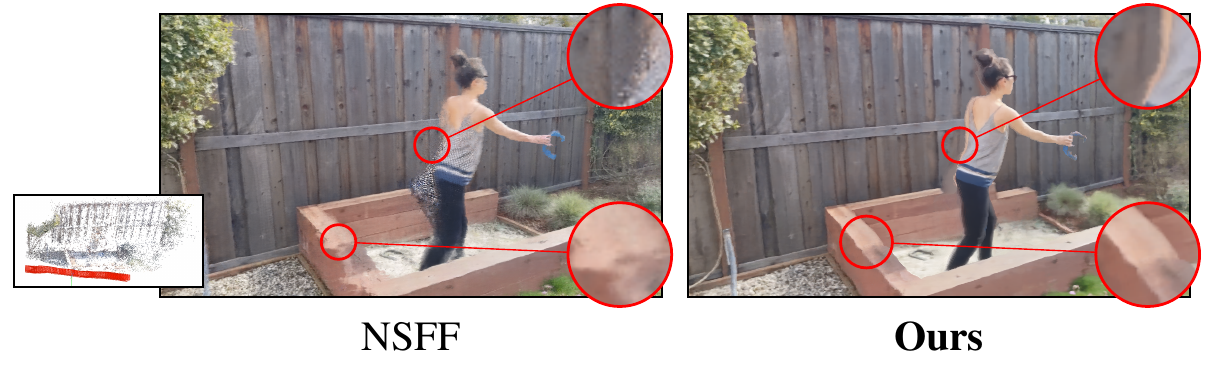}
	\vspace{-4mm}
	\caption{\small{Comparison with Li~\etal~\cite{flow-fields} on a challenging NVS case where we render the view on one end of the trajectory using the camera on the opposite end. The cameras are shown in the inset. Our method produces fewer artifacts in particular in the temporal occlusion regions.}}\label{fig:our_data}
\end{figure}

\noindent\textbf{Comparisons.}
The two most closely related methods are the method by Li~\etal~\cite{flow-fields} and that of Tretschk~\etal~\cite{nr-nerf}.
Both of these methods are pre-prints and we compare against them to offer context to the reader.

Table~\ref{table:ablation} shows a numerical comparison against these two methods, and two more~\cite{shih20203d,consistentdetph} whose numbers were reported by Li~\etal~\cite{flow-fields}.
The numbers show that we perform significantly better than the first three methods in the table.
We perform roughly on par with Li~\etal, with their method performing slightly better overall, and ours performing slightly better on the dynamic regions.
This is consistent with the fact that our main contribution is a global representation of the motion of the dynamic regions of the scene.
We offer a per-sequence breakdown of the comparison with the two most closely related methods in Table~\ref{table:breakdown}.

Figure~\ref{fig:results_with_insets} shows visual comparisons with Li~\etal and Tretschk~\etal.
Note how our method produces sharper results in many of the dynamic regions. We also show a comparison with the method by Li~\etal on two additional sequences in Figure~\ref{fig:our_data}.
These are two challenging cases in which the rendered view is at the location of the camera at one end of the capture sequence, but the values are from the time of the camera at the opposite end of the sequence.
Thanks to the fact that we model long-term trajectories, we can better reconstruct the scene.

Another approach for NVS of dynamic scenes is the method by Yoon~\etal~\cite{globalcoherent}.
Note that their method requires a mask for the dynamic regions to be given.
As their code is not publicly available we compare against their work by selecting a few frames from their own videos and synthesize a reasonably close view in space and time with our method.
Their results are overall sharper, which is consistent with the general observation that NeRF-based methods can be a bit blurrier.
However, several artifacts can be seen in their results around depth discontinuity regions.

\begin{figure}
	\setlength\tabcolsep{0.5pt}
	\centering
	\begin{tabular}{cc}
		\includegraphics[width=0.45\linewidth]{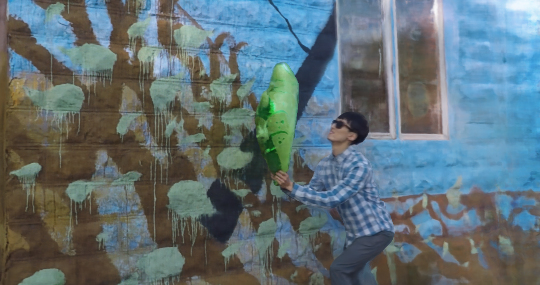} &
		\includegraphics[width=0.45\linewidth]{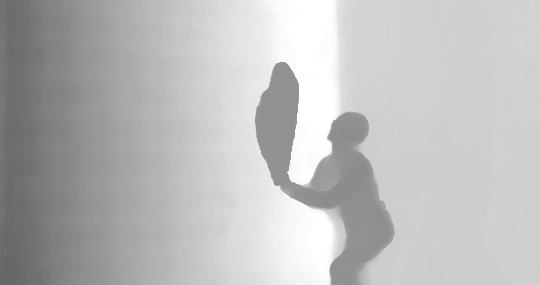} \\
		\includegraphics[width=0.45\linewidth]{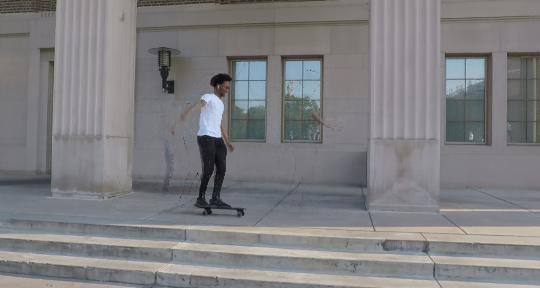} &
		\includegraphics[width=0.45\linewidth]{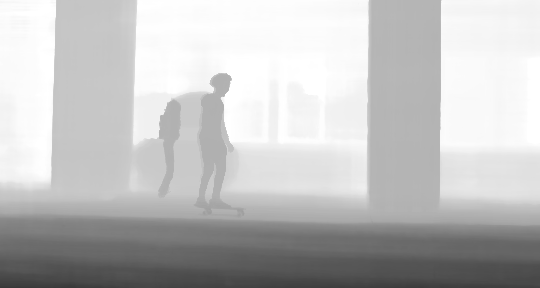} \\
	\end{tabular}
	\vspace{-3mm}
	\caption{\small{Failure cases for our method. Top: confuse background as moving region. Bottom: artifacts due to fast motion.}}\label{fig:failure}
\end{figure}

\noindent\textbf{Limitations.} Dynamic NVS from single camera is a very challenging task, and the quality of our result is sensitive to the context and motion of the input scene. Figure~\ref{fig:failure} shows two typical failure cases for our method. Our method may confuse specularity as dynamic motion and have trouble to track very thin objects with fast motion. Our method also shares the same limitations as other NeRF-based approaches: it requires long training per scene ($\approx$12hrs on 2 RTX2080 GPUs for a short sequence with 24 frames) and has a limited ability to recover high frequency dynamic details. As with others, we are also susceptible to degenerate cases of camera / object motion.

Some of the images we use in the paper are courtesy of Li~\etal~\cite{flow-fields} and the Nvidia Dynamic Scene Dataset~\cite{globalcoherent}.

\begin{table}
	\scriptsize
	\centering
	\begin{tabular}{l|cccc}
		\hline
		\multirow{2}{*}{Method}          & \multicolumn{2}{c}{Full scene} & \multicolumn{2}{c}{Dynamic parts only}                                            \\
		                                 & SSIM ($\uparrow$)              & LPIPS ($\downarrow$)                   & SSIM ($\uparrow$) & LPIPS ($\downarrow$) \\
		\hline
		3D Photo~\cite{shih20203d}       & 0.614                          & 0.215                                  & 0.486             & 0.217                \\
		Luo~\etal~\cite{consistentdetph} & 0.746                          & 0.141                                  & 0.530             & 0.207                \\
		NR-Nerf~\cite{nr-nerf}           & 0.526                          & 0.307                                  & 0.40              & 0.400                \\
		NSFF~\cite{flow-fields}          & 0.928                          & 0.045                                  & 0.758             & 0.097                \\
		\hline
		Ours w/o static                  & 0.885                          & 0.077                                  & 0.701             & 0.092                \\
		- local only                     & 0.878                          & 0.087                                  & 0.689             & 0.103                \\
		- w/o $p_\text{occ}$             & 0.889                          & 0.078                                  & 0.707             & 0.097                \\
		- w/o $\mathcal{L}^\text{cycle}$ & 0.881                          & 0.082                                  & 0.718             & 0.103                \\
		- w/o $\mathcal{L}^\text{svs}$   & 0.885                          & 0.082                                  & 0.711             & 0.096                \\
		- w/o $\mathcal{L}^\text{traj}$  & 0.879                          & 0.085                                  & 0.701             & 0.101                \\
		- w/o $\mathcal{L}^\text{depth}$ & 0.892                          & 0.081                                  & 0.697             & 0.108                \\
		\hline
		Ours (w static)                  & 0.915                          & 0.049                                  & 0.704             & 0.089                \\
		\hline
	\end{tabular}
	\vspace{-3mm}
	\caption{\small{Numerical results and ablation study for our method. We perform better than existing methods and
			on par with the concurrent method by Li~\etal\cite{flow-fields}.}}\label{table:ablation}
\end{table}

\begin{table}
	\footnotesize
	\centering
	\begin{tabular}{l|ccc}
		\hline
		\multirow{2}{*}{Sequence} & NR-NeRF~\cite{nr-nerf} & NSFF~\cite{flow-fields} & Ours                           \\
		                          & full / dyn.            & full / dyn.             & full / dyn.                    \\
		\hline
		Balloon1                  & 0.278~/0.398           & 0.062/0.175             & \textbf{0.051/~0.146}          \\
		Balloon2                  & 0.301~/0.395           & \textbf{0.030/0.065}    & 0.054/~0.074                   \\
		DynamicFace               & 0.164~/0.175           & 0.025/0.034             & \textbf{0.021}/~\textbf{0.022} \\
		Jumping                   & 0.415~/0.518           & \textbf{0.046/0.111}    & 0.056/~0.116                   \\
		Playground                & 0.363~/0.467           & 0.066/0.124             & \textbf{0.055/~0.092}          \\
		Skating                   & 0.349~/0.575           & \textbf{0.017/0.061}    & 0.036/~0.088                   \\
		Truck                     & 0.268~/0.346           & 0.026/0.052             & \textbf{0.025/~0.036}          \\
		Umbrella                  & 0.315~/0.322           & \textbf{0.089}/0.157    & 0.091/~\textbf{0.134}          \\
		\hline
	\end{tabular}
	\caption{\small{Breakdown of LPIPS metric  per scene from the Nvidia Dynamic Scene Dataset.}}\label{table:breakdown}
\end{table}

\begin{figure}
	\captionsetup[subfigure]{labelformat=empty}
	\centering

	\newlength{\heightyoon}
	\settoheight{\heightyoon}{\includegraphics[width=0.5\linewidth]{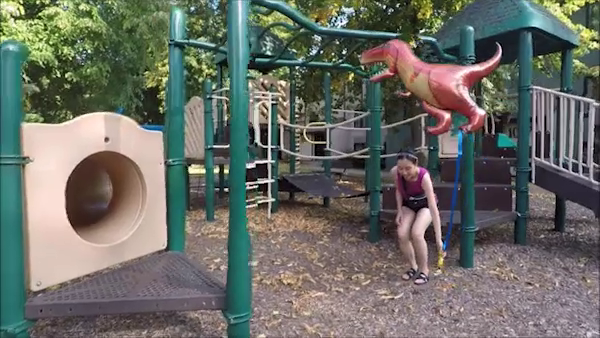}}

	\vspace{-4mm}
	\subfloat{\includegraphics[width=0.5\linewidth]{figures/compare_Jaeshin/playground_jaeshin.png}}~
	\subfloat{\includegraphics[width=0.5\linewidth, height=\heightyoon, trim={26 1 52 21}, clip]{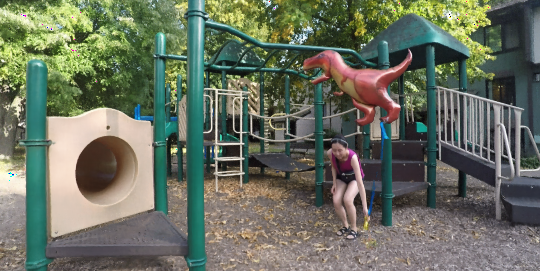}}\\
	\vspace{-3mm}
	\subfloat{\includegraphics[width=0.5\linewidth]{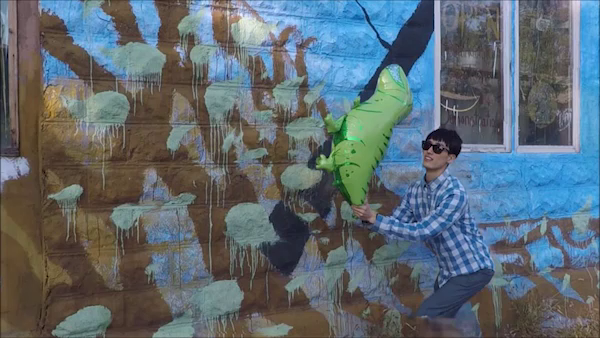}}~
	\subfloat{\includegraphics[width=0.5\linewidth, height=\heightyoon, trim={55 1 10 5}, clip]{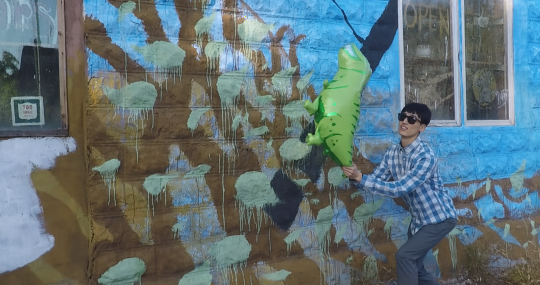}}\\
	\vspace{-3mm}
	\subfloat[Yoon~\etal~\cite{globalcoherent}]{\includegraphics[width=0.5\linewidth]{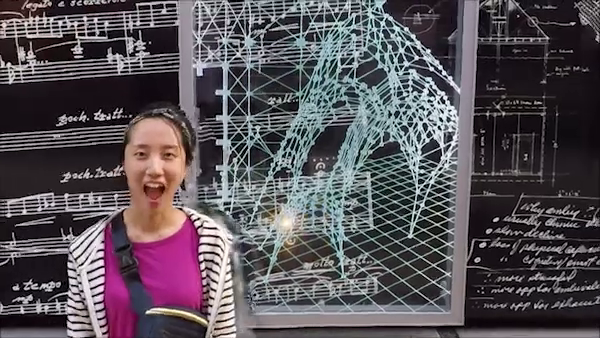}}~
	\subfloat[\bf{Ours}]{\includegraphics[width=0.5\linewidth, height=\heightyoon, trim={48 1 3 0}, clip]{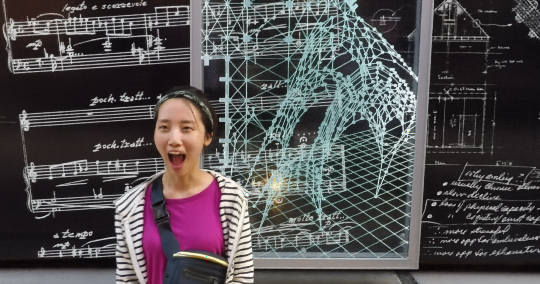}}
	\vspace{-3mm}
	\caption{\small{The code from Yoon~\etal is not available. We estimate the location and time of a frame from their results and synthesize it with our method. Note the artifacts at disocclusion regions, \eg, the front-most green pole in the top row, or the right side of the woman in the bottom.}}
\end{figure}

%% file: conclusions.tex

We present DCT-NeRF, a new coordinate-based neural representation that can render photorealistic novel views of dynamic scenes.
Our core contribution is the design of a dense trajectory field representation that captures the motion of dynamic parts of the scene over the entire input sequence. Using a  parametrization based on the discrete cosine transform (DCT), our trajectories are continuous and stable across the entire sequence. Further, we show how one can perform photometrically consistent sampling across arbitrary points in time by combining these parametric trajectory estimates with disocclusion prediction, and we use this technique to leverage information from any input image for any output image, ultimately producing compelling renditions of dynamic scenes.

%% file: supp.tex
\section{Implementation details}
 We use 10 frequencies for the spatial-temporal positional encoding, and use 4 frequencies for the directional and temporal input to the color output layer. Both the radiance field and the DCT trajectories are defined in normalized device coordinates (NDC). We linearly sample 128 depth values along the rays when performing volumetric rendering. Due to the intense computational cost of the current method, we do not perform additional depth sampling and train an extra ``fine'' network as done by Mildenhall~\etal~\cite{Mildenhall20eccv_nerf}.
\subsection{Regularization details}
\noindent\textbf{$\mathcal{L}^\text{cycle}$ details.} Given a point $\P_{t_0}$ at time $t_0$, by query the DCT trajectory $\traj_{\P_{t_0}}^{t_0}$ outputted by $\Psi(\P_{t_0}, t_0)$, we get its correspondences at time $t_1$, \ie $\P_{t_1} = \traj_{\P_{t_0}}^{t_0}(t_1)$. Denote the occupancy, color and DCT coefficients associated to the points $\P_{t_i}$ as $(\sigma_{t_i}, \c_{t_i}, \bvarphi_{t_i})$, and with particular note that
$\c_{t_0}=\c[\P_{t_0}, t_0](t_0)$ and $\c_{t_1}=\c[\P_{t_1}, t_1](t_0)$, the cycle consistency loss is evaluated as:
\begin{multline}
  \mathcal{L}^\text{cycle} = (1-p_\text{occ})(\|\bvarphi_{t_0}-\bvarphi_{t_1}\|_1 + 0.1\|\sigma_{t_0}-\sigma_{t_1}\|_1 \\ + 0.1\|\c_{t_0}-\c_{t_1}\|_1),
\end{multline}
where $p_\text{occ}$ is the probability of $\P_{t_0}$ being disoccluded by $\P_{t_1}$ as an empty region.

\noindent\textbf{$\mathcal{L}^\text{traj}$ details.} Given neighboring points $\P_{t_0}$ and $\P_{t_0}'$ (converted from NDC to Euclidean coordinates) along a ray at time $t_0$, the trajectory regularization loss is evaluated as:
\begin{equation}
  \begin{aligned}
    \mathcal{L}^\text{traj} & = \underbrace{ \|(\P_{t_0}-\P_{t_0\pm 1}) - (\P_{t_0}'-\P_{t_0\pm 1}')\|_1   }_{\text{spatial smoothness of scene flow.}}    \\
                            & + \underbrace{(1-p_\text{e})(\|(\P_{t_0}-\P_{t_0})' - (\P_{t_1}-\P_{t_1}')\|_1) }_{\text{as-rigid-as-possible.}}\\
                            & + \underbrace{ \|(\P_{t_0}-\P_{t_0\pm 1}) - (\P_{t_0}'-\P_{t_0\pm 1}')\|_1  }_{\text{temporal smoothness of trajectory / small scene flow.}}.
  \end{aligned}
\end{equation}
\noindent\textbf{Hyperparameters.} The full regularization loss is linearly combined as:
\begin{equation}
  \mathcal{L}^\text{cycle} + 0.1\mathcal{L}^\text{traj} + w_\text{SVS}\mathcal{L}^\text{SVS} + w_\text{depth}\mathcal{L}^\text{depth} + w_\text{flow}\mathcal{L}^\text{flow},
\end{equation}
where the initial $w_\text{depth}$ and $w_\text{flow}$ are set as 0.04 and 0.02, and then decreased by a factor of 0.1 every 7 epochs. $w_\text{SVS}$ is increased by a factor of 10 every 7 epochs from the initial value of $1e-5$, until it reaches $1e-2$.
\subsection{Optimization details}
We use Adam optimizer with default parameters, and set the learning rate as $5e-4$ for the first 70 epochs, then decrease the learning rate by 0.1 for the next 10 epochs for fine-tuning. For the temporal local-to-global procedure, we increase the temporal radius by 2 times every 10 epochs.\\
We use a batch size of $1024+512$ during training, with the first 1024 pixels uniformly sampled from training images, while the last 512 pixels are sampled uniformly only from the regions marked by a motion segmentation method. Sampling extra pixels from moving regions is also done by Li~\etal~\cite{flow-fields}, and we note that this is helpful to speed up convergence for small non-rigid regions, but not crucial to the end result of our method.

\section{Visual comparison for the ablation study}
\begin{figure*}
  \includegraphics[width=\linewidth]{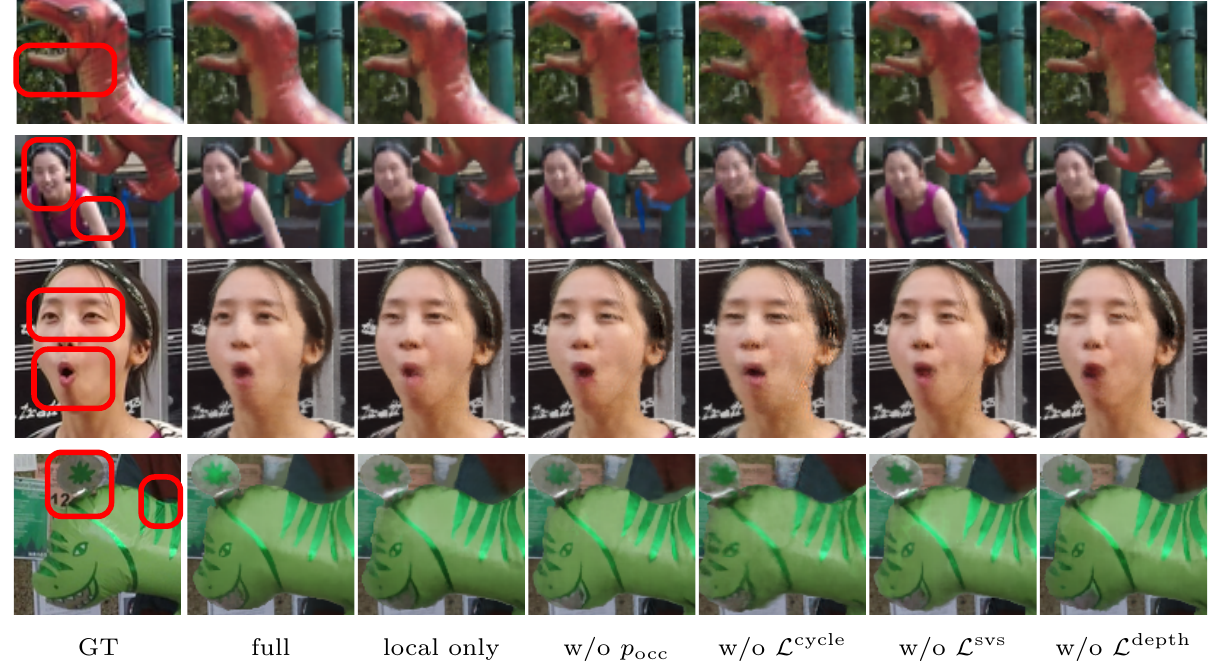}
  \caption{Visual comparison for ablation studies. The regions which show major difference between different versions of the method are marked with red rectangles.}
  \label{fig:compare_ab}
\end{figure*}

The visual comparison of different versions of the method for the ablation study is provided in Figure~\ref{fig:compare_ab}.

